\documentclass{egpubl}
\usepackage{pg2023}

\CGFStandardLicense

\usepackage[T1]{fontenc}
\usepackage{dfadobe}  

\usepackage{cite}  
\BibtexOrBiblatex
\electronicVersion
\PrintedOrElectronic
\ifpdf \usepackage[pdftex]{graphicx} \pdfcompresslevel=9
\else \usepackage[dvips]{graphicx} \fi

\usepackage{egweblnk} 

\usepackage{booktabs}
\usepackage{multirow}
\usepackage{amssymb}
\usepackage{float}
\usepackage{amsmath}
\usepackage{xcolor}
\usepackage{stfloats}
\usepackage[normalem]{ulem}

\title{Reconstructing 3D Human Pose from RGB-D Data with Occlusions}

\author[Bowen Dang, Xi Zhao, Bowen Zhang, He Wang]
{\parbox{\textwidth}{\centering 
    Bowen Dang$^{1}$\orcid{0000-0001-6065-6894}
    Xi Zhao\thanks{Corresponding Author}$^{1}$\orcid{0000-0002-3993-9870}
    Bowen Zhang$^{1}$\orcid{0000-0002-7919-1748}
    He Wang$^{2}$\orcid{0000-0002-2281-5679}}\\
{\parbox{\textwidth}{\centering
    $^1$Xi'an Jiaotong University, China\\ 
    $^2$University College London, United Kingdom}}
}

\newcommand{\del}[1]{{}}

\newcommand{\add}[1]{{#1}}

\newcommand{\deln}[1]{{}}

\newcommand{\addn}[1]{{#1}}

\newcommand{\addauthor}[1]{{#1 et al.}}

\newcommand{\metricname}[1]{{penetration}}

\begin{document}

\teaser{
 \includegraphics[width=\linewidth]{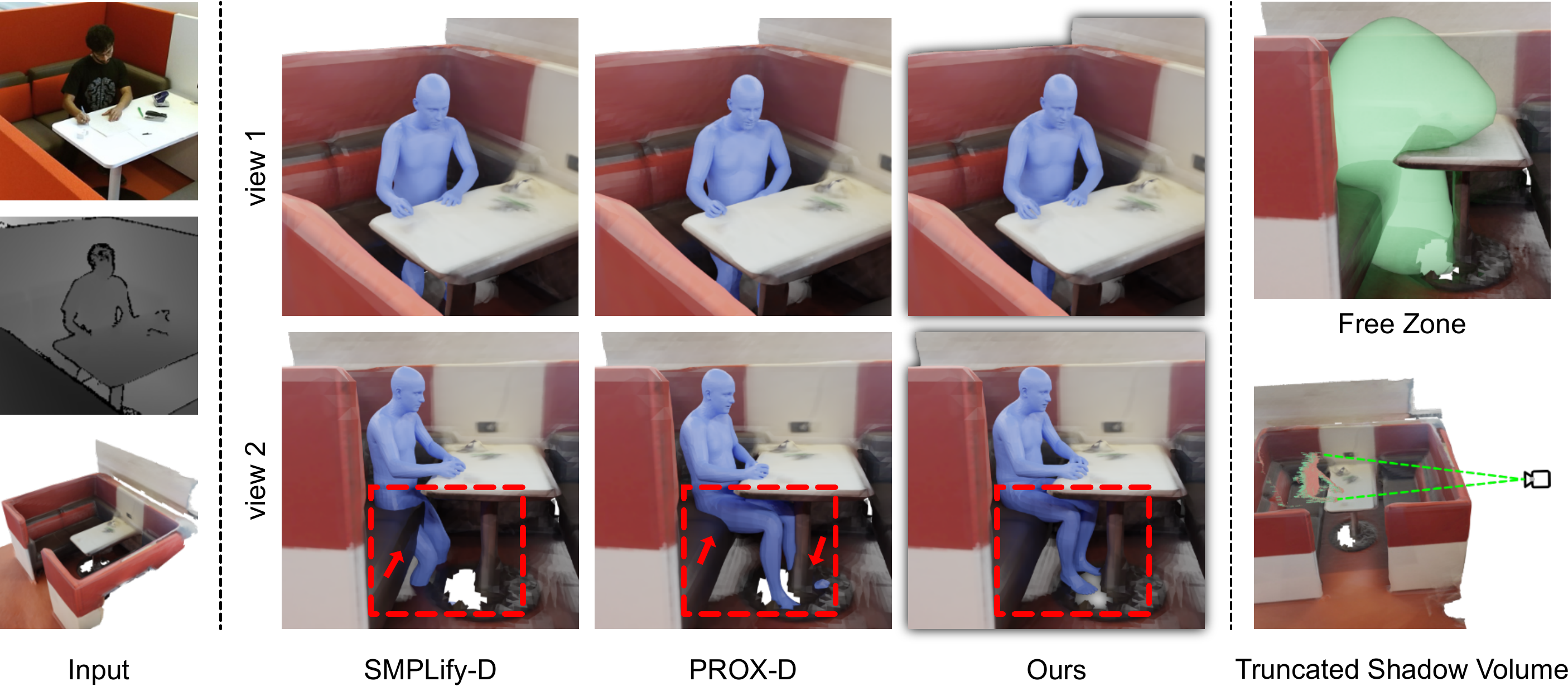}
 \centering
 \caption{Given a monocular RGB-D image and the scene mesh, our method reconstructs the 3D human body with a more plausible pose compared with SMPLify-D and PROX-D \cite{PROX} by reducing the solution space using free zone and truncated shadow volume. The difference between results is highlighted using red dashed box and arrows.}
 \label{figure:teaser}
}

\maketitle
\begin{abstract}
We propose a new method to reconstruct the 3D human body from RGB-D images with occlusions. The foremost challenge is the incompleteness of the RGB-D data due to occlusions between the body and the environment, leading to implausible reconstructions that suffer from severe human-scene penetration. To reconstruct a semantically and physically plausible human body, we propose to reduce the solution space based on scene information and prior knowledge. Our key idea is to constrain the solution space of the human body by considering the occluded body parts and visible body parts separately: modeling all plausible poses where the occluded body parts do not penetrate the scene, and constraining the visible body parts using depth data. Specifically, the first component is realized by a neural network that estimates the candidate region named the "free zone", a region carved out of the open space within which it is safe to search for poses of the invisible body parts without concern for penetration. The second component constrains the visible body parts using the "truncated shadow volume" of the scanned body point cloud. Furthermore, we propose to use a volume matching strategy, which yields better performance than surface matching, to match the human body with the confined region. We conducted experiments on the PROX dataset, and the results demonstrate that our method produces more accurate and plausible results compared with other methods.


\end{abstract}  


\newcommand{\smplifyd}{ E_{\mathrm{SMPLify-D}} }
\newcommand{\proxd}{ E_{\mathrm{PROX-D}} }

\newcommand{\shape}{ \beta }
\newcommand{\pose}{ \theta }
\newcommand{\face}{ \psi }
\newcommand{\trans}{ t }

\newcommand{\jterm}{ E_J } 
\newcommand{\dterm}{ E_d }
\newcommand{\rterm}{ E_r }
\newcommand{\pterm}{ E_p }
\newcommand{\cterm}{ E_c }
\newcommand{\fzterm}{ E_{\mathrm{fz}} }
\newcommand{\svterm}{ E_{\mathrm{tsv}} }

\newcommand{\dweight}{ \lambda_d }
\newcommand{\rweight}{ \lambda_r }
\newcommand{\pweight}{ \lambda_p }
\newcommand{\cweight}{ \lambda_c }
\newcommand{\fzweight}{ \lambda_{\mathrm{fz}} }
\newcommand{\svweight}{ \lambda_{\mathrm{tsv}} }
\newcommand{\weight}{ \lambda_i }

\newcommand{\depth}{ P_b }
\newcommand{\scene}{ P_s }
\newcommand{\query}{ P_q }
\newcommand{\vertices}{ V_b }
\newcommand{\grid}{ P_{\mathrm{grid}} }
\newcommand{\fzpoints}{ P_{\mathrm{fz}} }
\newcommand{\svpoints}{ P_{\mathrm{tsv}} }
\newcommand{\internal}{ P_{\mathrm{int}} }

\newcommand{\pcamera}{ p_c }
\newcommand{\pstart}{ p_s }
\newcommand{\pend}{ p_e }
\newcommand{\dray}{ r }
\newcommand{\mdepth}{ d_l }
\newcommand{\idepth}{ d_i }

\newcommand{\fznet}{ \mathrm{FZNet} }
\newcommand{\bfield}{ F_b }
\newcommand{\sfield}{ F_s }
\newcommand{\bgt}{ \mathrm{GT}_b }
\newcommand{\sgt}{ \mathrm{GT}_s }
\newcommand{\clamp}{ \delta }
\newcommand{\fzthre}{ \mu }
\newcommand{\rootjoint}{ J_{\mathrm{root}} }

\newcommand{\vfront}{ v_f }
\newcommand{\vback}{ v_b }
\newcommand{\ifront}{ i_f }
\newcommand{\iback}{ i_b }
\newcommand{\fzef}{ \rho_{\mathrm{fz}} }
\newcommand{\svef}{ \rho_{\mathrm{tsv}} }

\newcommand{\faces}{ F_b }
\section{Introduction}

3D human reconstruction is an important research area with broad applications in human behavior understanding and human-scene interaction analysis \cite{LEMO}. Most current works focus on reconstructing 3D human body from monocular RGB images and matching the reconstructed result with the 2D image \cite{SMPLify, HMR, SMPL-X, SPIN, EFT, METRO, LVD}. With the development of human-scene interaction datasets \cite{PiGraphs, PROX, GPA, EgoBody}, many methods \cite{PROX, LEMO, MocapDeform} have been dedicated to reconstructing 3D human body that not only matches with the 2D human contour in image but also keeps reasonable spatial relationship with the environment by using 3D scene information. In this paper, we also follow this line and work on 3D human reconstruction from monocular RGB-D images. We specifically focus on situations which contain close interactions and serious occlusions between the human and the environment. 


3D human reconstruction methods can be divided into two types: optimization-based and regression-based methods. Optimization-based methods attempt to reconstruct the 3D human body by minimizing an objective function to optimize the parameters of the human body model \cite{SMPLify, SMPL-X} or vertices \cite{LVD}. Regression-based methods directly regress the parameters of the human body model \cite{HMR} or vertices \cite{METRO} in an end-to-end manner, which requires a lot of data to train and might lead to inaccurate pose. So the regression-based method is normally followed by a post-process to optimize the pose \cite{SPIN, EFT}. \deln{Current methods suffer from penetrations between the body and the environment, although they all consider the penetration terms \mbox{\cite{PROX}}.}\addn{Although Current methods all consider the penetration terms \cite{PROX}, they suffer from penetrations between the body and the environment.} The most important reason is that it is quite hard to improve or resolve the penetration once it happens and the system \deln{sticks into}\deln{may stuck in}\addn{may stick in} the local minimum. The penetration problem is getting even harder when dealing with scenes with close interactions and serious occlusions.

Our method falls into the optimization type. Our key idea is to explicitly reduce the solution space based on scene information and prior knowledge. We propose two strategies to constrain the solution space. First, with the 3D scene around the human body, we can infer the possible region where the human body can lie in without penetrations. We refer to this region as the \textit{free zone (FZ)}. Considering a person sitting on a chair with her/his legs under a table in front of the chair, the free zone is mainly the space between the chair and the table. With the free zone, we can significantly reduce the solution space for searching plausible poses of the invisible body parts. Second, inspired by the concept of shadow volume in computer graphics \cite{ShadowVolume}, we consider the camera as a point light source, and construct a \textit{truncated shadow volume (TSV)} to constrain the possible space for the visible body parts. The main idea is that the partially scanned body should fit and locate "behind" the seen point cloud in the direction of camera ray. In summary, in stead of estimating the human pose inside the whole space, we use above two strategies to make a confined region, within which the the 3D human pose is searched. By doing this, our results can avoid most penetrations between the body and the environment. 



We design two methods \deln{to realize the}\addn{based on} above strategies. First, inspired by neural implicit fields \cite{DeepSDF, OccNet, UDF, GraspingField, CHORE}, we use a neural network to estimate the free zone. \del{Given a randomly sampled point in the whole space, the network can predict two distances: the distance from the point to the body and that to the scene. We construct the free zone by collecting those points whose distance to the body is below a certain threshold.}\add{Given a randomly sampled point in the whole space, the network can predict two field values: the body field value and scene field value. We construct the free zone by collecting those points whose body field value is below a certain threshold.} Second, we compute the shadow volume of the body point cloud by shooting rays from the camera toward each point in the scanned body point cloud until they hit the environment. We further truncate the shadow rays from the scanned body point cloud to a certain maximum length limit and build the truncated shadow volume. We represent the truncated shadow volume discretely by uniformly sampled points along the truncated shadow rays. After obtaining the free zone and the truncated shadow volume, the next step is to match the body with them. We sample the human body by a differentiable interpolation algorithm, which produces points inside the body. Then we minimize the distances from the points in the body to points in the confined region. Our experiments show that such a volume matching method outperforms the surface matching method in terms of accuracy and robustness. 


We also propose a more comprehensive metric to evaluate the \metricname{}. Non-Collision (NC) is a traditional metric that measures the \metricname{} between the body and the environment. It calculates the ratio of body vertices with positive Signed Distance Field (SDF) values. However, NC only considers body surface vertices and does not work well when part of the environment is inside the body. To address this limitation, we introduce a Volume Non-Collision (VNC) metric, which considers the points inside the body when calculating the similar ratio.

In summary, our contributions are as follows:
\begin{enumerate}
    \item We propose two novel schemes, which consider the invisible and visible body part separately, to reduce the solution space for optimization-based pose reconstruction systems;
    \item We design novel methods to \deln{realize and apply the}\addn{apply} above strategies by matching the body with the confined region as a volume;
    \item We demonstrate that our system can reconstruct human poses with higher accuracy and less penetration compared to baseline methods.
\end{enumerate}





\section{Related Work}

\noindent\textbf{\del{Optimization-based}3D Human Reconstruction}: \add{3D human reconstruction methods can be divided into optimization-based and regression-based methods.} Optimization-based \del{3D human reconstruction}methods aim to reconstruct the 3D human body that matches with the RGB or RGB-D image by iteratively optimizing the parameters of the human body model \cite{SMPLify, SMPL-X, PROX, LEMO, MocapDeform}. The key difference of these methods lie in the objective function, which typically consists of two parts: data terms and regularization terms \cite{HMRSurvey}. Data terms are designed to align the human body with the input data, including RGB data and depth data. Regularization terms are used to constrain the parameters and prevent unrealistic poses and shapes. \deln{SMPLify}\addauthor{Bogo} \cite{SMPLify} \addn{propose to} iteratively fit the SMPL \cite{SMPL} human body model to the 2D keypoints detected by DeepCut \cite{DeepCut}. \deln{SMPLify-X}\addauthor{Pavlakos} \cite{SMPL-X} follow a similar scheme but provide more detailed output for hands and face. However, these methods often suffer from visual artifacts, including scene penetration, feet sliding, and body leaning \cite{HMRSurvey}. To address these limitations, recent research has focused on leveraging scene information to constrain the pose and produce more plausible results. \deln{PROX}\addauthor{Hassan} \cite{PROX} propose a human-scene penetration term and a contact term \deln{on the top of SMPLify-X}\addn{on top of SMPLify-X \cite{SMPL-X}}. By considering the scene constraint, \deln{PROX}\addn{they} achieve more realistic results with less penetration and necessary contact. Given partial observations, \deln{LEMO}\addauthor{Zhang} \cite{LEMO} introduce a motion smoothness prior to address jittering issues and employ a contact-aware motion infiller to infer plausible motions of occluded body parts. Compared with PROX-D \cite{PROX}, \deln{LEMO}\addn{they} produce improved results with smoother motions and more plausible body-scene interactions. In contrast to methods that treat the scene as a rigid object, \deln{MocapDeform}\addauthor{Li} \cite{MocapDeform} jointly optimize the human body and the non-rigid deformation of the scene, leading to superior accuracy in reconstructing the 3D human body compared with other methods. \add{Regression-based methods directly regress the parameters of the human body model in an end-to-end manner \cite{HMR}. \deln{HMR}\addauthor{Kanazawa} \cite{HMR} design a deep neural network to predict the parameters of SMPL human body model without requiring the 3D paired data. \deln{SPIN}\addauthor{Kolotouros} \cite{SPIN} incorporate a post-process optimization module based on HMR \cite{HMR} to improve the precision of the result.} Our method is based on the optimization backbone, and it considers two new constraints: the free zone term and truncated shadow volume term to reduce the solution space.

\noindent\textbf{Neural Implicit Field}: The 3D model can be represented explicitly or implicitly. Recently, many works have been focused on using the implicit functions such as the DeepSDF \cite{DeepSDF}, Occupancy Networks \cite{OccNet}, and UDF \cite{UDF} to represent the 3D shape. There are also works using this method to represent the relationship between two objects \cite{GraspingField, CHORE}. \deln{Grasping Field}\addauthor{Karunratanakul} \cite{GraspingField} represent the hand and the grasped object using implicit field including the signed distances to the hand and the object. This representation can be used for grasp generation. \deln{CHORE}\addauthor{Xie} \cite{CHORE} \addn{propose to} extract the the body distance field, object distance field, object pose field, and body part field from an RGB image. These fields are \deln{utilized}\addn{used} to optimize the parameters of the human body and the object, facilitating accurate and realistic reconstruction. Our method leverages a similar approach to \deln{Grasping Field and CHORE in modeling}\addn{model} the relationship between the human body and the scene.  


\noindent\textbf{Interaction Representation}: Various methods have been proposed to extract the interaction feature between two parts, such as a hand and a object or a human body and the surrounding environment \cite{IBS, RaG}. \deln{IBS}\addauthor{Zhao} \cite{IBS} propose to extract the Interaction Bisector Surface \addn{(IBS)} between two objects using a geometry-based method, and use this feature for the classification and retrieval of 3D objects. \addauthor{She} \cite{RaG} use IBS to represent the gripper-object interaction between gripper and object to solve the high-DOF reaching-and-grasping problem. The study of interaction feature between the human body and the surrounding scene has been explored in the field of 3D human reconstruction and generation \cite{PSI, PLACE, POSA}. \deln{PSI}\addauthor{Zhang} \cite{PSI} use the conditional Variational Auto-Encoder (cVAE) \cite{cVAE} to predict semantically plausible 3D human body based on latent scene features. The generated human body are further refined by incorporating scene constraints to ensure plausible interactions. \deln{PLACE}\addauthor{Zhang} \cite{PLACE} model the proximal relationship between the human body and the scene using BPS \cite{BPS} feature. \deln{POSA}\addauthor{Hassan} \cite{POSA} propose a body-centric human-scene interaction model that can be generalized to new scenes. These approaches contribute to the understanding and representation of human-scene interactions in the context of 3D human reconstruction and generation. Different from above methods that constrain the human pose in latent space, we explicitly reduce the solution space, within which the human pose is searched to align with the image and avoid penetrations.

\section{Method}

\begin{figure*}[htbp]
  \centering
  \mbox{}
  \hfill
  \includegraphics[width=\linewidth]{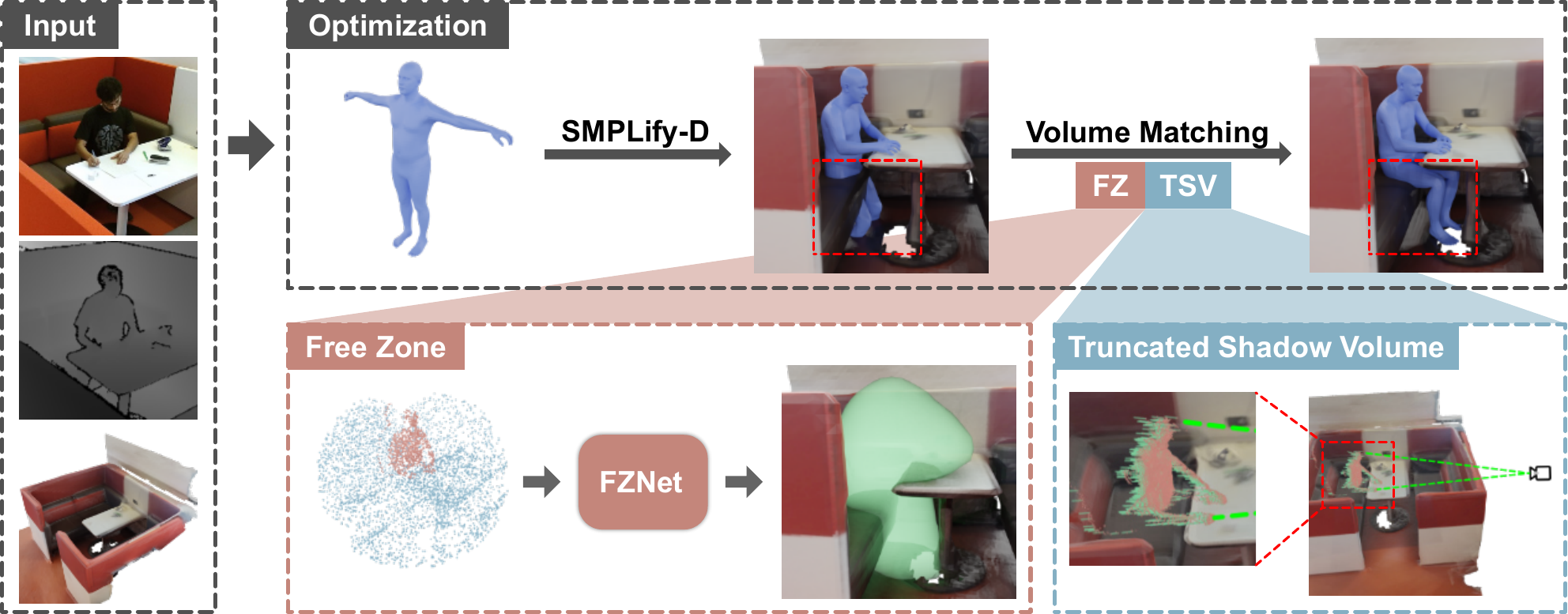}
  \hfill
  \mbox{}
  \caption{The overview of our method. The input include a monocular RGB-D image and the scene mesh. In the first stage, we optimize the SMPL-X parameters from T-pose using SMPLify-D to get an initial result. In the second stage, we employ two strategies to reduce the solution space and \deln{utilize}\addn{use} a volume matching algorithm to match the human body with the confined region. We design the free zone network (FZNet) to estimate the region where the human body can be positioned, which is used to constrain the invisible body parts; we calculate the truncated shadow volume behind the scanned body point cloud and use it to constrain the visible body parts.}
  \label{figure:overview}
\end{figure*}

Our goal is to reconstruct the human body mesh from a monocular RGB-D image and the scene mesh. Our main idea is to reduce the solution space for body parts based on their visibility. In this section, we present details of our method.

\subsection{Preliminaries}

\noindent\textbf{SMPL-X human body model}: The SMPL-X \cite{SMPL-X} human body model is a differentiable function used to model the human body. It takes shape parameters $\shape \in \mathbb{R}^{12}$, pose parameters $\pose \in \mathbb{R}^{K \times 3}$, facial expression parameters $\face$, and global translation $\trans \in \mathbb{R}^3$ as input. The output is a human body mesh $M_b=(\vertices,\faces)$ composed of \deln{$N_b=10475$}\addn{$N_b$} vertices\deln{$\vertices$and triangular faces $\faces$}:

\begin{equation}
    M(\shape,\pose,\face,\trans): \mathbb{R}^{|\shape| \times |\pose| \times |\face| \times |\trans|} \rightarrow \mathbb{R}^{N_b \times 3}
\end{equation}

\noindent The pose parameters include poses for the body, hands, and jaw with axis-angle representation. $K$ represents the total number of joints in the model, including 22 for the body, 30 for the hands (15 per hand), and 3 for the face. $J(\shape)$ is the 3D coordinates of each joint, which can be inferred from the vertices of the human body mesh using linear blend skinning. The parameters of the SMPL-X human body model can be optimized by adjusting the coordinates of the vertices or joints.

\noindent\textbf{SMPLify-D and PROX-D}: \deln{In SMPLify-D and PROX-D \mbox{\cite{PROX}}, the human body is reconstructed to align with the RGB-D data by optimizing the parameters of the SMPL-X human body model.}\addn{SMPLify-D and PROX-D \cite{PROX} reconstruct the human body to align with the RGB-D data by optimizing the parameters of the SMPL-X human body model.} The objective function \addn{of SMPLify-D} is represented as follows:

\begin{equation}
    \smplifyd = \jterm + \dweight \dterm + \rweight \rterm
\end{equation}

\noindent $\jterm$ is a re-projection loss that aims to minimize the robust weighted distance between 2D joints estimated from the RGB image using OpenPose \cite{OpenPose} and the 2D projections of the corresponding posed 3D joints of SMPL-X human body model. The depth term $\dterm$ minimizes the distances between the visible body vertices $\vertices^v \in \vertices$ and scanned body point cloud $\depth$\addn{ which is extracted using the depth image and the body segmentation mask detected by DeepLab V3 \cite{DeepLabV3} from the RGB image}. \deln{The scanned body point cloud $\depth$ is extracted using the depth image and the body segmentation mask detected by DeepLab V3 \mbox{\cite{DeepLabV3}} from the RGB image.}The regularization term $\rterm$ is composed of multiple terms including pose prior term, shape prior term, self-penetration term, et al. \add{Specifically, the self-penetration term \cite{SMPL-X} is used to avoid collisions between different body parts.} $\dweight$ and $\rweight$ denote the weights for the depth term and regularization term respectively. PROX-D extends the SMPLify-D framework by adding a penetration term $\pterm$ and a contact term $\cterm$ to enforce scene constraint on the human body:




\begin{equation}
    \proxd = \smplifyd + \pweight \pterm + \cweight \cterm
\end{equation}

\noindent The penetration term $\pterm$ penalizes all penetrating vertices using the Signed Distance Field (SDF) of the scene mesh $M_s$. The contact term $\cterm$ encourages contact and proximity between the body and the scene by minimizing the distances from body vertices to scene vertices around contact areas. $\pweight$ and $\cweight$ denote the weights for the penetration term and contact term respectively.


\subsection{Overview}

Our optimization framework takes a monocular RGB-D image and the scene mesh as input. The output is the human body mesh. As depicted in Figure \ref{figure:overview}, our optimization framework consists of two stages. In the first stage, we use SMPLify-D to obtain an initial result. In the second stage, we employ two strategies to reduce the solution space and \deln{utilize}\addn{use} the volume matching algorithm to match the human body with the confined region. The objective function of our method is defined as follows:

\begin{equation}
    E = \smplifyd + \fzweight \fzterm + \svweight \svterm + \cweight \cterm
\end{equation}

\noindent We extend SMPLify-D by introducing the free zone term $\fzterm$ and truncated shadow volume term $\svterm$ into the objective function. To prevent the issue of the body floating and encourage necessary contact, we also incorporate the contact term \deln{borrowed}from PROX-D. $\fzweight$ and $\svweight$ denote the weights for the free zone term and truncated shadow volume respectively.


\begin{figure}[t]
  \centering
  \includegraphics[width=\linewidth]{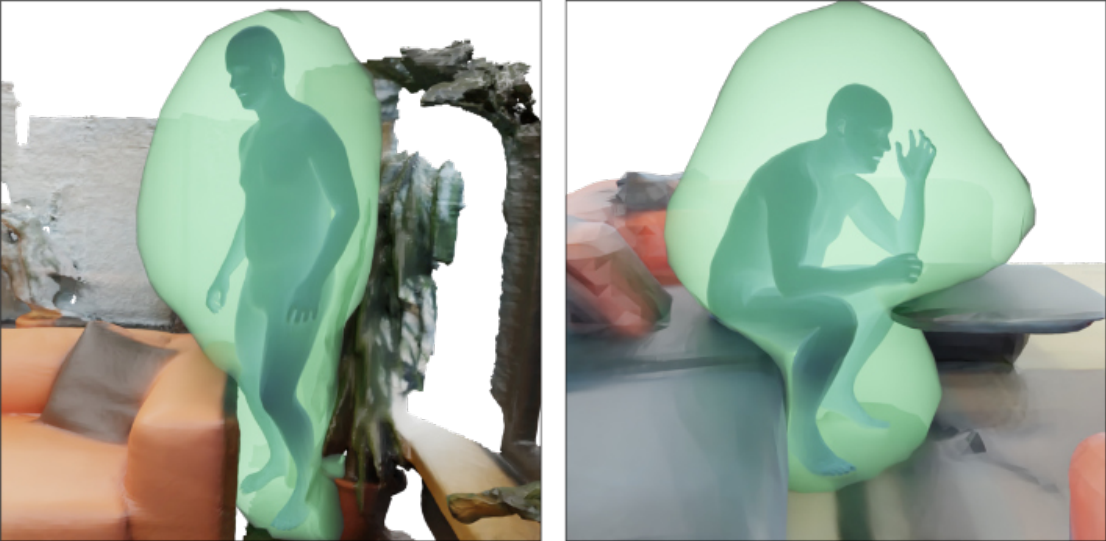}
  \caption{Examples of the free zone and the corresponding reconstructed body.}
  \label{figure:fz_example}
\end{figure}

\subsection{Free Zone}

\begin{figure}[t]
  \centering
  \includegraphics[width=\linewidth]{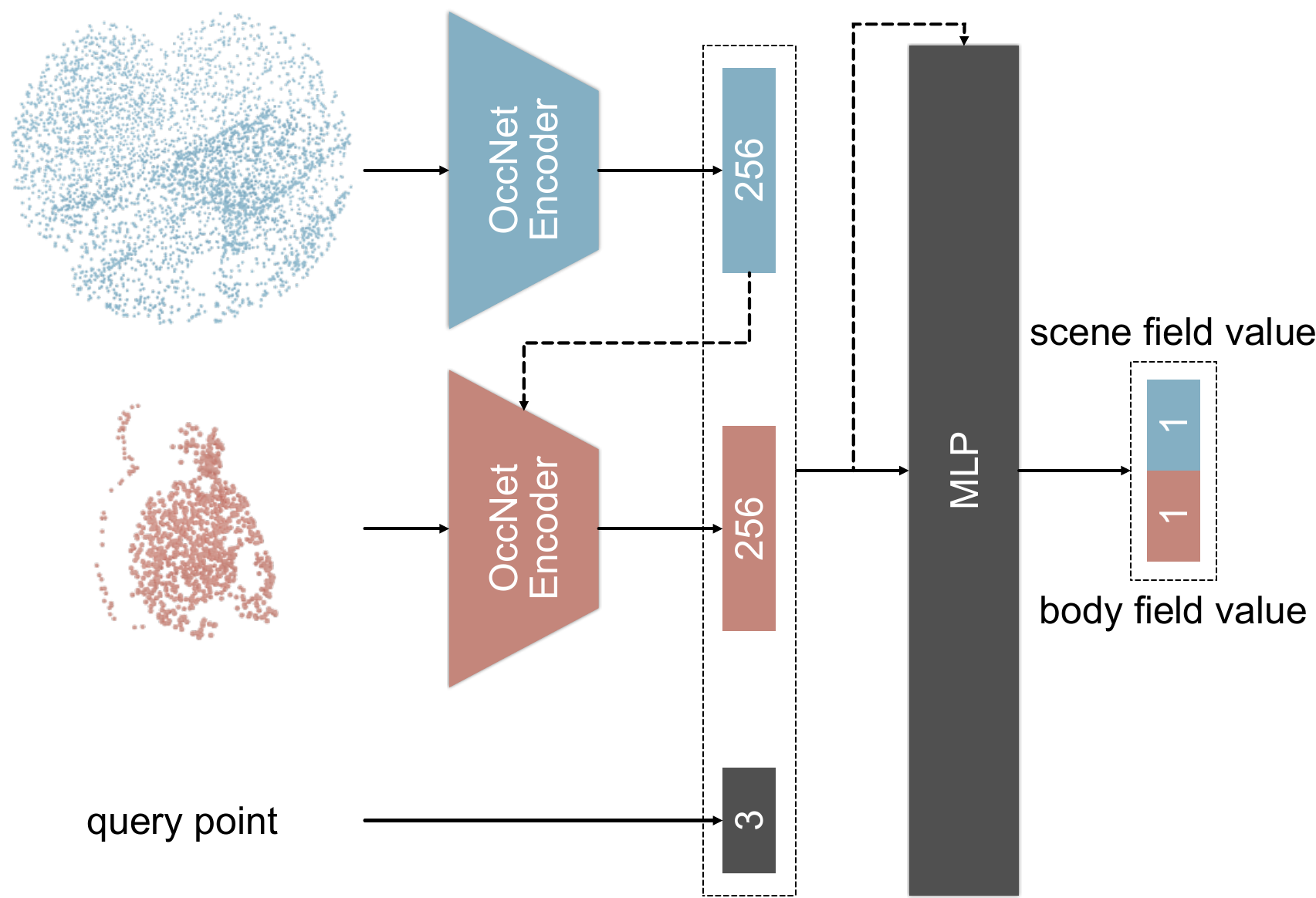}
  \caption{The structure of the free zone network (FZNet). The input are the scene point cloud, the scanned body point cloud, and a query point. \del{The output are the distances from the query point to the body and the scene.}\add{The output are the body field value and scene field value.}}
  \label{figure:fz_network}
\end{figure}

Current methods tries to solve the penetration problem by penalizing all penetrating body vertices using the SDF of the scene \cite{PROX}. The effectiveness of this intuitive approach heavily relies on the accuracy and completeness of the scene. In reality, limitations in the scanning devices and the complexity of the scene can cause errors. Besides, this approach becomes even more ineffective in scenarios where the body part penetrates into the scene deeply or penetrates through thin objects. To solve these problems, we introduce a novel approach that \deln{utilizes}\addn{uses} a neural network to learn the potential region where the human body can lie in without penetrating the environment. This region is referred to as the "free zone", and it serves as a confined region within which we search for plausible poses of the invisible body parts. In Figure \ref{figure:fz_example}, we show some examples of free zone and corresponding reconstructed body. 

We propose to use an encoder-decoder-style network, the free zone network (FZNet), to estimate the free zone. The structure of FZNet is shown in Figure \ref{figure:fz_network}. The scanned body point cloud $\depth$ and scene point cloud $\scene$ are encoded separately using the \deln{encoder from Occupancy Networks}\addn{OccNet encoder} \cite{OccNet}. The encoded scene point cloud feature is fed back to the body point cloud encoder to enforce condition. Then the scene feature, the body feature and the query point $p$ are concatenated together and fed to a MLP decoder to predict the\deln{the} field value for $p$ in body field $\bfield$ and scene field $\sfield$:

\begin{equation}
    \fznet(\depth, \scene, p) = (\bfield(p), \sfield(p))
\end{equation}

\noindent \del{Specifically, $\depth$ is extracted using the depth image and the body segmentation mask detected by DeepLab V3 from the RGB image. We use Farthest Point Sampling (FPS) to obtain 1024 points.}\add{To extract $\depth$, we first detect the body segmentation mask from the RGB image using DeepLab V3 \cite{DeepLabV3}. Then, we obtain the whole point cloud, which includes both the human and the scene, from the depth image. Finally, we extract the body point cloud from the whole point cloud using the segmentation mask and obtain 1024 points using Farthest Point Sampling (FPS).} $\scene$ is sampled on the scene mesh with 4096 points. We train the free zone network by minimizing the $L_1$ distance between clamped prediction and ground truth distance: 

\begin{equation}
\begin{aligned}
    L_{\fznet} = \sum_{p \in \query} (& | \mathrm{min}(\bfield(p), \clamp) - \mathrm{min}(\bgt(p), \clamp) | + \\
    & | \mathrm{min}(\sfield(p), \clamp) - \mathrm{min}(\sgt(p), \clamp) |)
\end{aligned}
\end{equation}

\noindent where $\query$ are the sampled query points. $\clamp$ is the clamping distance and we set it as 0.1.


After getting the initial pose, we \deln{utilize}\addn{use} the \deln{pre-trained}\addn{trained} free zone network to get the free zone points. Assuming that the root joint of current human body is $\rootjoint$. First, we sample scene point cloud within an unit sphere from scene mesh $M_s$ centered at $\rootjoint$. Next, we uniformly sample query points $\grid$ within a voxel grid with side length of 2 and resolution of 64 centered at $\rootjoint$. For each sampled query point, we query its body field value. We then retain points with field value below a certain threshold, and consider these points as the free zone points used in the second stage of the optimization:


\begin{equation}
    \fzpoints = \{p|\bfield(p) < \fzthre, p \in \grid\}
\end{equation}

\noindent where $\fzthre$ is the \del{distance}threshold and we set it as 1e-3.

\subsection{Truncated Shadow Volume}

\begin{figure}[t]
  \centering
  \includegraphics[width=\linewidth]{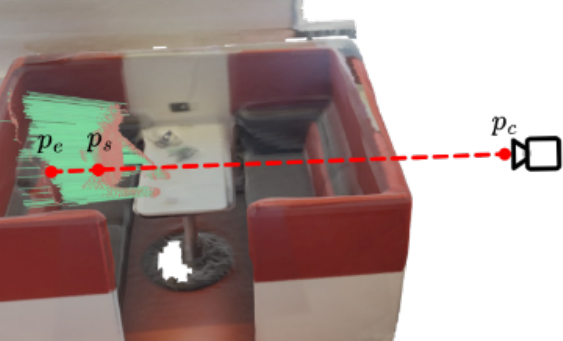}
  \caption{The illustration of calculating the shadow volume points from the scanned body point cloud. Pink points denote the scanned body point cloud and green points denote the shadow volume points. The maximum length limit for truncation is set to infinity for better visualization.}
  \label{figure:sv_example}
\end{figure}

Existing methods may generate implausible results that appear in front of the scanned body point cloud. However, the visible body parts should be located behind the scanned body point cloud in the direction of the camera ray. We propose to calculate the truncated shadow volume behind the scanned body point cloud and use it to constrain the visible body parts.

We represent the shadow volume by shadow rays that are cast from the camera toward the scanned body. In Figure \ref{figure:sv_example}, we show the shadow rays, from which we can compute the truncated shadow rays. Let $\pcamera$ represent the camera coordinate. The ray direction $\dray$ is calculated as:

\begin{equation}
    \dray = \frac{\pstart - \pcamera}{\parallel \pstart - \pcamera \parallel}
\end{equation}

\noindent where $\pstart \in \depth$ denotes the body point, which is also the start point of the truncated shadow rays. We set a maximum length limit $\mdepth$ to truncate the shadow rays, ensuring that the truncated shadow ray is not far from the scanned body. The distance from the first intersection point of the ray with the scene mesh to the start point is $\idepth$. Then the end point $\pend$ is computed by:

\begin{equation}
    \pend = \pstart + \min(\mdepth, \idepth) \cdot \dray
\end{equation}

\noindent By adding offsets to the start point $\pstart$ at a fixed interval, we can obtain a set of intermediate points that gradually move toward the end point $\pend$. These intermediate points form the truncated shadow volume points $\svpoints$. We empirically set the maximum length limit $\mdepth$ as 0.1 and the interval as 0.01.

\subsection{Differentiable Volume Matching}


We propose to represent the human body as a volume composed of points located inside the body. To implement this, we need to obtain the internal points from body vertices while ensuring the process is differentiable.

\begin{figure}[t]
  \centering
  \includegraphics[width=\linewidth]{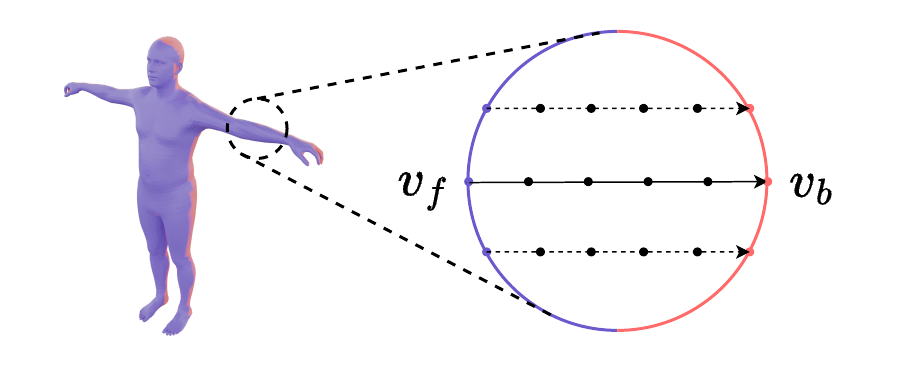}
  \caption{The process of getting the internal points from body vertices. Purple denotes the front and red denotes the back.}
  \label{figure:volume}
\end{figure}

As illustrated in Figure \ref{figure:volume}, we use an interpolation-based method to get the internal points from vertices. We use the SMPL-X human body model at T-pose as a template mesh to get the interpolation vertex pairs. We first divide the body vertices into front vertices and back vertices by calculating the visibility using a virtual camera in front of the body. Then we sample front vertices using FPS to make the vertices uniformly distributed. For each front vertex $\vfront$, we cast a ray towards the back vertices. We can get the nearest back vertex $\vback$ from the intersection point of the ray with the human body mesh. $\vfront$ and $\vback$ form an interpolation vertex pair. By repeating above steps for all selected front vertices, we can obtain a set of interpolation vertex pairs. Then we perform linear interpolation and get 6 points for each pair, resulting in the internal points $\internal$. Assuming the body is locally convex, $\internal$  will not extend beyond the body boundary and can remain the body pose and shape. 




Free zone can be seen as a superset of the body and truncated shadow volume can be seen as a subset of the body. We match the human body with free zone points and truncated shadow volume points separately. For the free zone points, we employ the following loss:

\begin{equation}
    \fzterm = \sum\limits_{p_i \in \internal} \fzef(\min\limits_{p_j \in \fzpoints} \parallel p_i - p_j \parallel)
\end{equation}

\noindent where $\fzef$ denotes a robust Geman-McClure error function\cite{GMEF} for down weighting the points in $\fzpoints$ that are far from $\internal$, so that the human body will not become too fat. For the truncated shadow volume points, we employ the following loss:

\begin{equation}
    \svterm =  \sum\limits_{p_i \in \svpoints}  \svef(\min\limits_{p_j \in \internal} \parallel p_i - p_j \parallel)
\end{equation}

\noindent where $\svef$ denotes a robust Geman-McClure error function for down weighting points in $\internal$ that are far from $\svpoints$, so that the human body will not become too thin.

\section{Experiments}


\subsection{Datasets}

We conduct experiments on the PROX dataset \cite{PROX}. The PROX dataset is divided into two parts: a qualitative set and a quantitative set. The qualitative set consists of monocular videos of 20 human subjects interacting with 12 indoor scenes. The dataset contains 100K RGB-D frames recorded at 30 fps, along with the scene mesh. The pseudo-ground-truth SMPL-X \cite{SMPL-X} parameters are fitted using PROX-D \cite{PROX}. The quantitative set consists of 180 static RGB-D frames, with one human subject wearing markers and interacting with a living room containing daily furniture. The ground-truth SMPL-X parameters are fitted using MoSh++ \cite{AMASS}. We \deln{utilize}\addn{use} the scene data from POSA \cite{POSA}, which aligns the scene data of the PROX dataset for easy processing.


\subsection{Experiment Details}

\textbf{Dataset Split}: We randomly split the qualitative set into training and testing sets with a ratio of 4:1. The training set is used to train the free zone network. We exclude the data that has high penetrations and extract frames every second. The testing set is used to evaluate our method. We also conduct experiments on the quantitative set. 

\noindent \textbf{Free Zone Network Training}: We sample 20K query points every training sample. Among these points, 95\% are located near the surface and 5\% are uniformly distributed within the bounding box. To generate points near the surface, we first sample surface points on both the body and the scene. Then we introduce perturbations to each surface point along three axes. These perturbations are generated by applying zero-mean Gaussian noise with variances of 0.02 and 0.002, resulting in two query points every surface point. For each generated point, we calculate its nearest distance to the body points and scene points. This allows us to determine the point's proximity to the body and scene, providing valuable information for training the free zone network. To enhance the generalization of the free zone network, we apply two data augmentation techniques, including random rotation along the z-axis and online query points generation. By employing these data augmentation techniques, we aim to create a diverse training set that enables the free zone network to generalize on different visible ratios and scene types. We train the free zone network using the Adam optimizer \cite{Adam} with a learning rate of 1e-4. We employ a step learning rate schedule with a decay rate of 0.5 after 100 epochs. The models are trained for 200 epochs on a single 3090Ti GPU.


\noindent \textbf{Optimization}: We use the scene as the world coordinate and both the free zone and truncated shadow volume are defined in this coordinate. Before matching with the confined region, we transform the human body from the camera coordinate to the world coordinate. The weights for each term in the objective function are set empirically. Specifically, we set the weight of the free free zone term to 5e2 and the weight of the truncated shadow volume term to 2e2. We use the same optimizer with SMPLify-X \cite{SMPL-X}.

\begin{figure}[t]
  \centering
  \includegraphics[width=\linewidth]{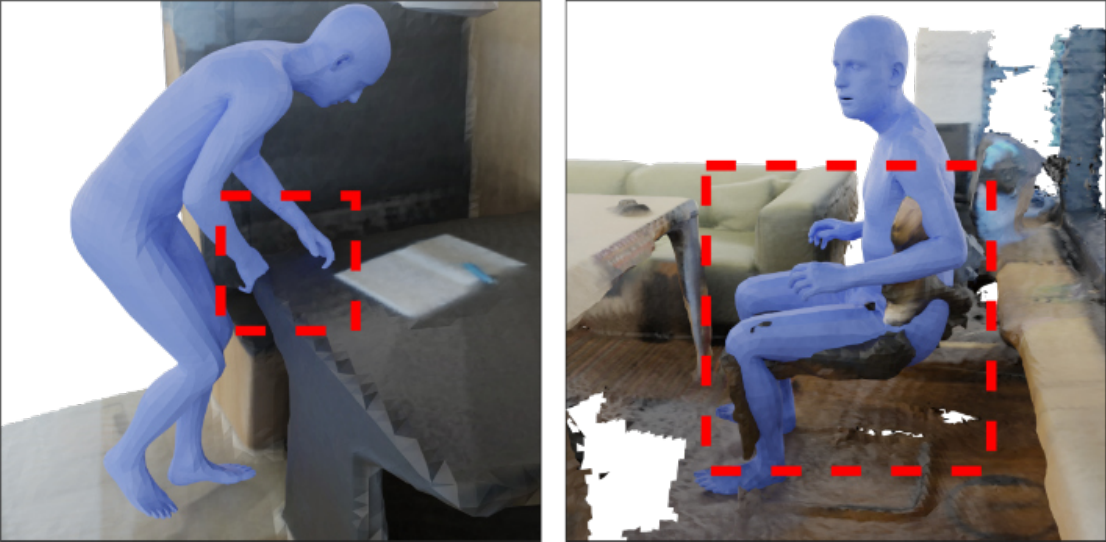}
  \caption{Examples to illustrate that Non-Collision can not evaluate the \metricname{} correctly. The penetration area is highlighted using red dashed box.}
  \label{figure:vnc}
\end{figure}

\subsection{Results}

\begin{figure*}[htbp]
  \centering
  \mbox{} 
  \hfill
  \includegraphics[width=0.77\linewidth]{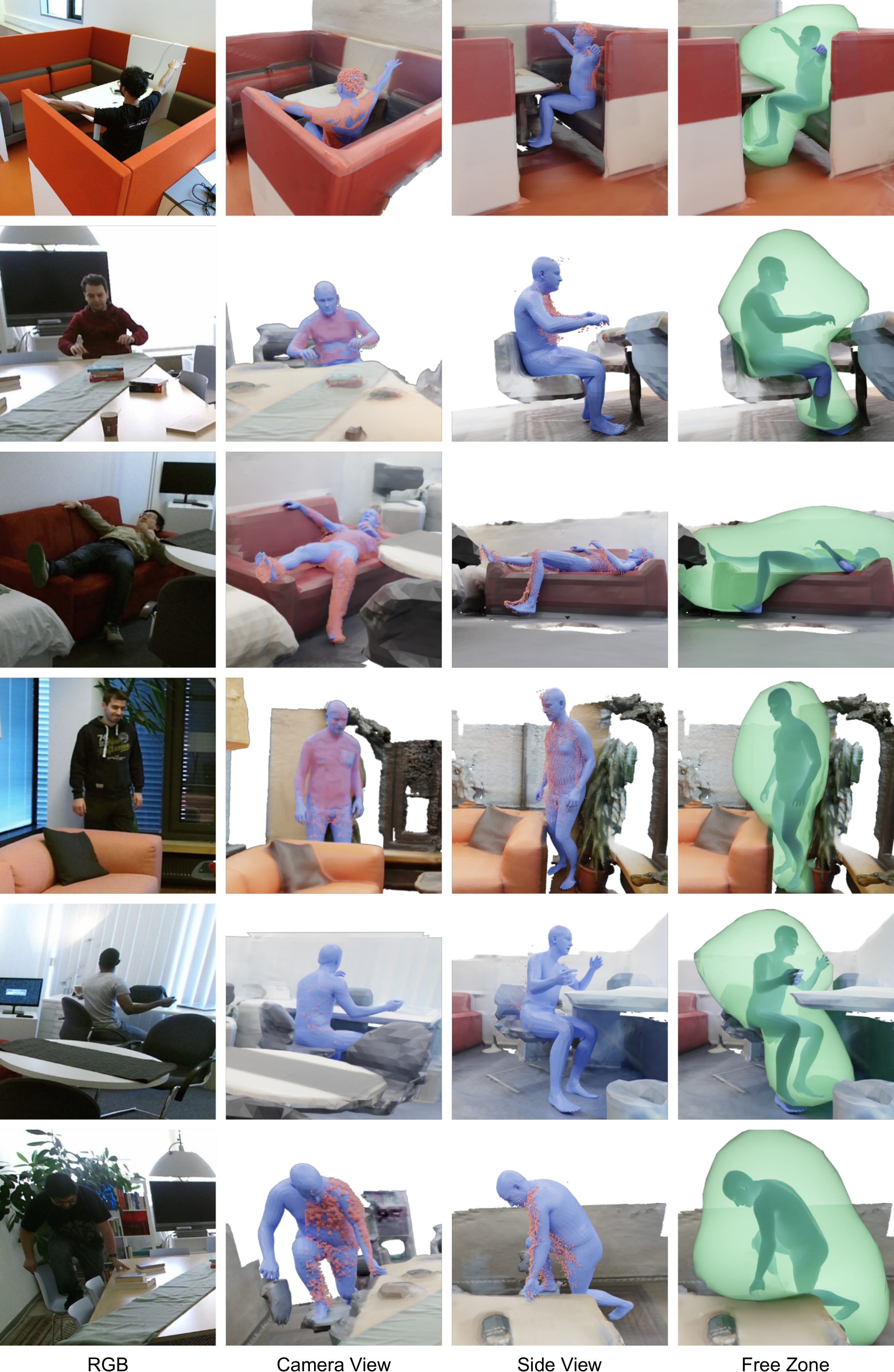}
  \hfill
  \mbox{}
  \caption{Gallery of our results. We show examples from different scenes and poses. Pink points denote the scanned body point cloud.}
  \label{figure:ours}
\end{figure*}

Figure \ref{figure:ours} presents a gallery of our results. We can see that our method can produce accurate and plausible poses on different scene types and visible ratios. When the human body is sitting on a chair and only the upper body is visible (row 1, 2), the free zone can guide the human body to be in a sitting pose, no matter the human is sitting back or facing to the camera. When the human has a lot of contact with the scene (row 3), such as when the human is lying on a sofa, our method can produce plausible results with little penetration and necessary contact. In cases where the scenes are more complex (row 4), such as when the human is standing between a sofa and a pot of plants, our free zone network can still identify the correct region. Even when the human is not captured by the depth camera (row 5), the free zone network can still estimate the free zone correctly using only the scene information. Additionally, when the human is in an irregular pose (row 6), our method can still produce results that match with the image. 


\subsection{Comparison and Evaluation}

We compare our method with SMPLify-D and PROX-D \cite{PROX} using a series of evaluation metrics on both PROX quantitative and qualitative sets.

\noindent \textbf{Evaluation Metrics}: We evaluate the performance using a set of metrics, which can be categorized into accuracy metrics and plausibility metrics. We assess the 3D reconstruction accuracy using Joint Position Error (JPE) and Vertex-to-Vertex Error (V2V). JPE measures the mean per joint position error and V2V represents the mean error between corresponding vertices of the reconstructed and the ground truth mesh. We also apply Procrustes alignment to the meshes and calculate the aligned JPE (p.JPE) and aligned V2V (p.V2V). To evaluate the alignment accuracy with the depth data, we \deln{utilize}\addn{use} a Partial Matching (PM) metric, which measures the mean distance between each body point and its closest vertex on the reconstructed mesh. The PM score is lower when the matching is more accurate.


To evaluate the plausibility, we employ the Non-Collision (NC) metric introduced in PSI \cite{PSI}. This metric is calculated as the ratio of body vertices with a positive SDF value. The NC score is lower when the penetration is more severe. However, we find this metric does not work well in scenarios where the environment is inside the body. For example, in Figure \ref{figure:vnc}, the NC scores of these two examples are both 0.93. However, the \metricname{}s are completely different. In the left example, only part of the hands penetrates the table. In the right example, the chair penetrates into the body completely. This highlights a potential discrepancy in evaluating \metricname{} using the NC. To address this problem, we propose a modified version called Volume Non-Collision (VNC), which calculates the ratio of the internal points with a positive SDF value. The VNC score is lower when the penetration is more severe. By considering the points inside the body, this metric provides a more comprehensive evaluation of \metricname{}. In Figure \ref{figure:vnc}, the VNC scores of these two examples are 0.99 and 0.79 respectively, suggesting a more accurate evaluation of \metricname{} compared with the NC. 

\noindent\textbf{Quantitative Results}: The comparison results on the quantitative set are listed in Table \ref{table:quantitative}. Our method achieves the lowest error in all metrics, although there is only one scene in the quantitative set and the interactions are simple.


\begin{table}[htb]
  \centering
  \small
  \caption{Results on PROX quantitative set.}
  \begin{tabular}{l|cccc}
  \toprule
    & JPE $\downarrow$ & V2V $\downarrow$ & p.JPE $\downarrow$ & p.V2V $\downarrow$ \\
  \midrule
    SMPLify-D & 73.80 & 76.81 & 45.61 & 44.57 \\
    PROX-D & 69.46 & 72.70 & 42.43 & 42.20 \\
    \textbf{Ours} & \textbf{66.74} & \textbf{70.04} & \textbf{41.86} & \textbf{41.46} \\
  \bottomrule
  \end{tabular} 
  \label{table:quantitative}
\end{table}

\noindent \textbf{Qualitative Results}: The comparison results on the qualitative set are listed in Table \ref{table:qualitative}. Our method achieves the best performance in all metrics.

\begin{table}[htb]
    \centering
    \small
    \caption{Results on PROX qualitative set.}
    \resizebox{\linewidth}{!}{
    \begin{tabular}{l|cccc|ccc}
    \toprule
        & $\pterm$ & $\cterm$ & $\fzterm$ & $\svterm$ & NC $\uparrow$ & VNC $\uparrow$ & PM/1e-3 $\downarrow$\\
    \midrule
        SMPLify-D & & & & & 95.6\% & 95.2\% & 4.42 \\
        PROX-D & \checkmark & \checkmark & & & 96.8\% & 97.3\% & 4.53 \\
        \textbf{Ours} & & \checkmark & \checkmark & \checkmark & \textbf{97.1\%} & \textbf{97.9\%} & \textbf{4.10} \\
    \bottomrule
    \end{tabular}}
    \label{table:qualitative}
\end{table}

\begin{figure*}[htbp]
  \centering
  \mbox{} 
  \hfill
  \includegraphics[width=0.77\linewidth]{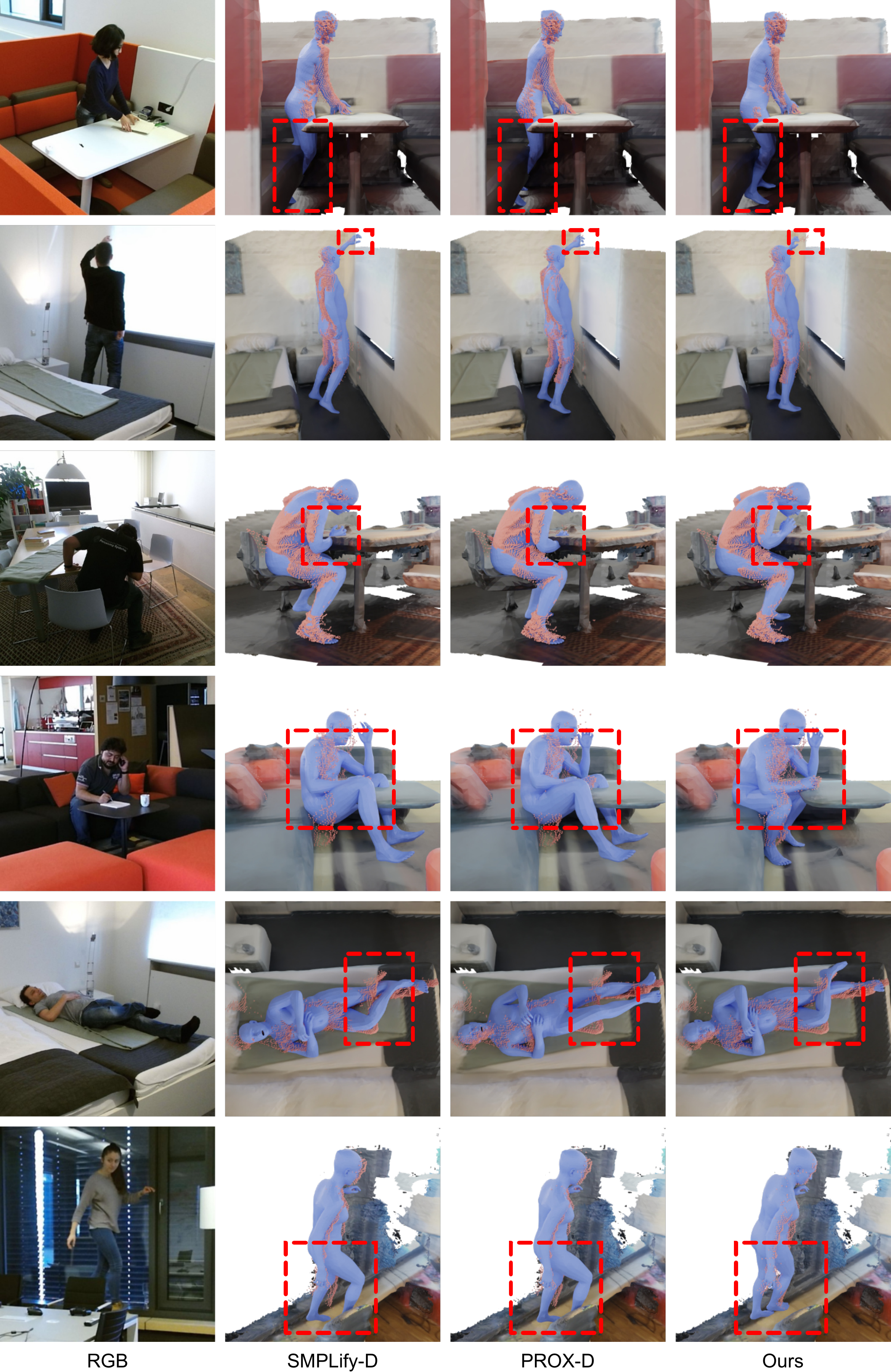}
  \hfill
  \mbox{}
  \caption{Comparison of the reconstructed results by our method with those of SMPLify-D and PROX-D. Pink points denote the scanned body point cloud. The difference between results is highlighted using red dashed box.}
  \label{figure:qualitative}
\end{figure*}

In Figure \ref{figure:qualitative}, we visually compare our method with other methods on the PROX qualitative set. In the first 4 rows, where the human is partially occluded by the scene or by themselves, both SMPLify-D and PROX-D produce results where some parts penetrate the scene. However, our method can infer the correct pose of the invisible body parts and avoid penetrations. When some body parts penetrate deeply into objects, such as a leg penetrating into a sofa or a hand penetrating into a wall (row 1, 2), it is hard for current methods to pull the body out of the object completely. However, our method \deln{utilizes}\addn{uses} the free zone to guide the body away from the object, effectively reducing the penetration. Our method can also handle cases where some body parts penetrate through thin objects like a table (row 3, 4), preventing such penetrations from occurring. In the last 2 rows, where different body parts overlap with each other, our results exhibit better alignment with the scanned body point cloud compared with other methods thanks to the constraint of the truncated shadow volume.

We further analyse of the qualitative results by scene types and visible ratios. Results based on different scene types are shown in Table \ref{table:qualitive_scene}. We can see that our method has a more significant improvement on scenes with sitting booths or beds. In scenes with sitting booths, the legs may be occluded by the table or chair. Current methods often \deln{reconstructs}\addn{reconstruct} the wrong pose with the leg penetrating the chair and struggles to pull the leg out. Our method is more capable of handling such occlusions with the aid of the free zone term. For scenes with beds, the geometry of the bed can serve as a strong clue to generate accurate free zone and truncated shadow volume, leading to more accurate and plausible reconstruction. Results based on different visible ratios are shown in Table \ref{table:qualitive_visible}. The visible ratio is calculated by comparing the scanned body point cloud with the reconstructed results of PROX-D. We can see that our method has a more significant improvement when the visible ratio is lower, which demonstrates the effectiveness of our method in situations with serious occlusions.

\begin{table*}[t]
    \centering
    \small
    \caption{\del{Qualitative results}\add{Evaluation} based on different scene types.}
    \resizebox{\linewidth}{!}{
    \begin{tabular}{l|ccc|ccc|ccc|ccc}
    \toprule
        & \multicolumn{3}{c|}{Sitting Booth} & \multicolumn{3}{c|}{Chair} & \multicolumn{3}{c|}{Sofa} & \multicolumn{3}{c}{Bed}\\
    \cmidrule{2-13}
        & NC $\uparrow$ & VNC $\uparrow$ & PM/1e-3 $\downarrow$ & NC $\uparrow$ & VNC $\uparrow$ & PM/1e-3 $\downarrow$ & NC $\uparrow$ & VNC $\uparrow$ & PM/1e-3 $\downarrow$ & NC $\uparrow$ & VNC $\uparrow$ & PM/1e-3 $\downarrow$\\
    \midrule
        SMPLify-D & 93.8\% & 93.1\% & 8.97 & 96.9\% & 96.5\% & 3.17 & 96.7\% & 96.2\% & 3.75 & 92.6\% & 92.3\% & 5.14 \\
        PROX-D & 95.8\% & 96.4\% & 9.20 & 97.4\% & 97.8\% & 3.38 & 97.5\% & 97.9\% & 4.14 & 95.5\% & 96.4\% & 5.41 \\
        \textbf{Ours} & \textbf{96.1\%} & \textbf{97.3\%} & \textbf{8.44} & \textbf{97.6\%} & \textbf{98.2\%} & \textbf{3.04} & \textbf{97.8\%} & \textbf{98.4\%} & \textbf{3.56} & \textbf{96.5\%} & \textbf{97.3\%} & \textbf{4.72} \\
    \bottomrule
    \end{tabular}}
    \label{table:qualitive_scene}
\end{table*}

\begin{table*}[t]
    \centering
    \small
    \caption{\del{Qualitative results}\add{Evaluation} based on different visible ratios.}
    \resizebox{\linewidth}{!}{
    \begin{tabular}{l|ccc|ccc|ccc|ccc}
    \toprule
        & \multicolumn{3}{c|}{$0\%-25\%$} & \multicolumn{3}{c|}{$25\%-50\%$} & \multicolumn{3}{c|}{$50\%-75\%$} & \multicolumn{3}{c}{$75\%-100\%$}\\
    \cmidrule{2-13}
        & NC $\uparrow$ & VNC $\uparrow$ & PM/1e-3 $\downarrow$ & NC $\uparrow$ & VNC $\uparrow$ & PM/1e-3 $\downarrow$ & NC $\uparrow$ & VNC $\uparrow$ & PM/1e-3 $\downarrow$ & NC $\uparrow$ & VNC $\uparrow$ & PM/1e-3 $\downarrow$\\
    \midrule
        SMPLify-D & 94.9\% & 91.1\% & 12.40 & 94.3\% & 93.8\% & 5.71 & 96.6\% & 96.7\% & 2.85 & \textbf{99.8\%} & \textbf{99.8\%} & 1.16 \\
        PROX-D & 97.1\% & 96.9\% & 14.62 & 95.9\% & 96.4\% & 5.97 & 97.4\% & 98.1\% & 2.69 & 99.3\% & 99.7\% & \textbf{1.12} \\
        \textbf{Ours} & \textbf{97.6\%} & \textbf{98.0\%} & \textbf{12.00} & \textbf{96.4\%} & \textbf{97.3\%} & \textbf{5.35} & \textbf{97.7\%} & \textbf{98.3\%} & \textbf{2.56} & 99.2\% & 99.7\% & 1.13 \\
    \bottomrule
    \end{tabular}}
    \label{table:qualitive_visible}
\end{table*}

\subsection{Ablation Study}

We conduct the ablation study on the PROX qualitative set and the results are shown in Table \ref{table:ablation}. We consider the following ablation versions:

\begin{itemize}
  \item Ours (w/o FZ): we test how our method performs when the free zone term is removed.
  \item Ours (w/o TSV): we test how our method performs when the truncated shadow volume term is removed. 
  \item Ours (w/o VM): we replace the volume matching with surface matching.
\end{itemize}

\begin{figure}[t]
  \centering
  \includegraphics[width=\linewidth]{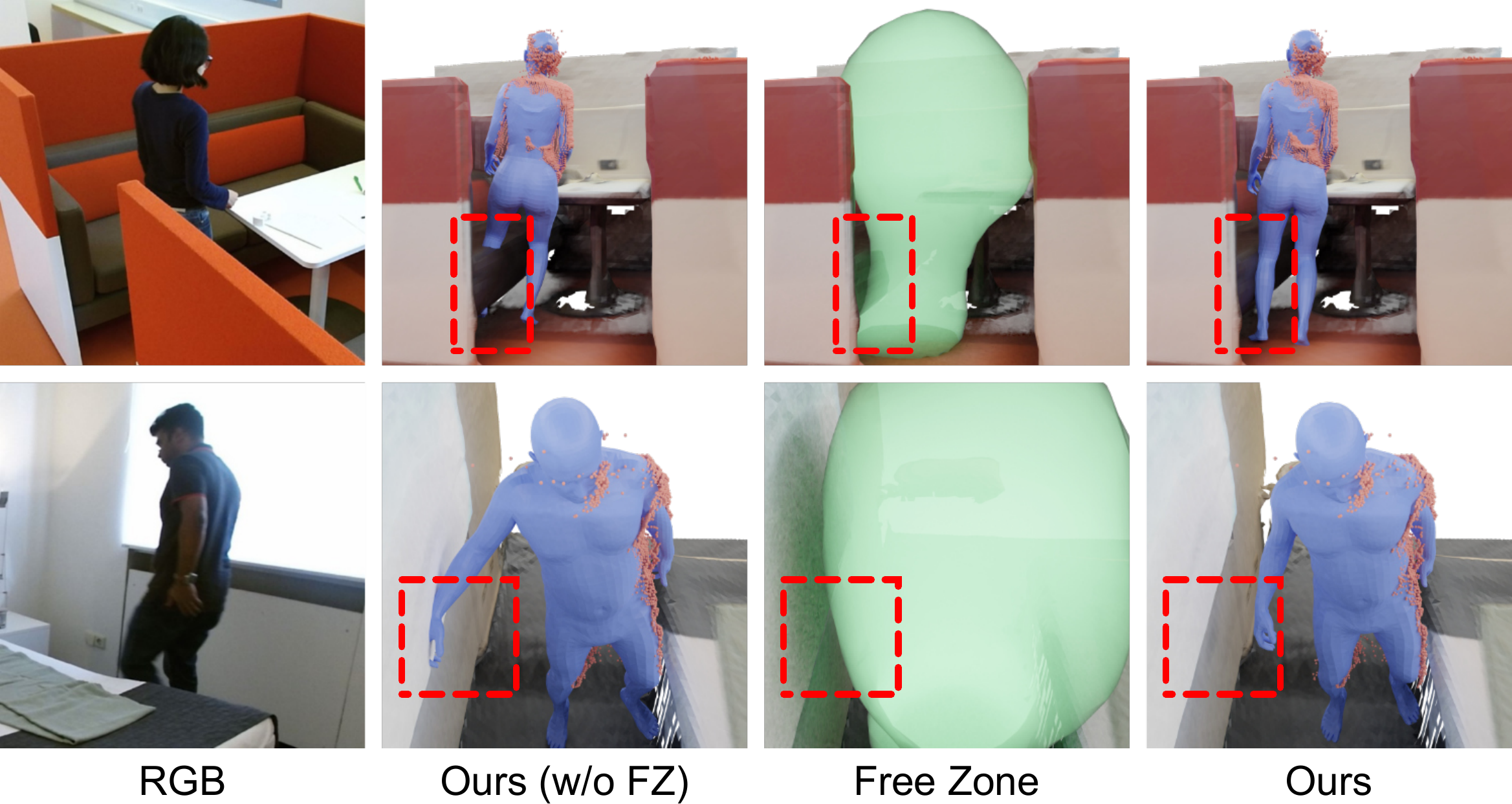}
  \caption{Ablation study: compare our method with the version that does not use free zone term. Pink points denote the scanned body point cloud. The difference between results is highlighted using red dashed box.}
  \label{figure:ablation_fz}
\end{figure}

\noindent We can see that the final version which includes all the proposed components achieves the best overall performance. 

\begin{table}[htb]
    \centering
    \small
    \caption{Ablation \deln{results}\addn{study} on PROX qualitative set.}
    \resizebox{\linewidth}{!}{
    \begin{tabular}{l|ccc|ccc}
    \toprule
        & $\fzterm$ & $\svterm$ & VM & NC $\uparrow$ & VNC $\uparrow$ & PM/1e-3 $\downarrow$ \\
    \midrule
        Ours (w/o FZ) & & \checkmark & \checkmark & 95.1\% & 95.4\% & 4.18 \\
        Ours (w/o TSV) & \checkmark & & \checkmark & 96.6\% & 96.4\% & 4.47  \\
        Ours (w/o VM) & \checkmark & \checkmark & & 96.2\% & 96.3\% & 4.25 \\
        \textbf{Ours} & \checkmark & \checkmark & \checkmark & \textbf{97.1\%} & \textbf{97.9\%} & \textbf{4.10} \\
    \bottomrule
    \end{tabular}}
    \label{table:ablation}
\end{table}

Without the free zone term, the reconstructed results have more penetrations with the scene, and both the NC score and VNC score decrease. Figure \ref{figure:ablation_fz} presents two representative examples to demonstrate this effect. In the first example, when our method is applied without the free zone term, the reconstructed result shows that the left leg penetrates into the sitting booth. In the second example, we can observe a similar scenario where the right hand penetrates into the wall. By using the free zone term to constrain the invisible body parts, we can effectively reduce the penetration issue, ensuring a more plausible reconstruction.

\begin{figure}[t]
  \centering
  \includegraphics[width=\linewidth]{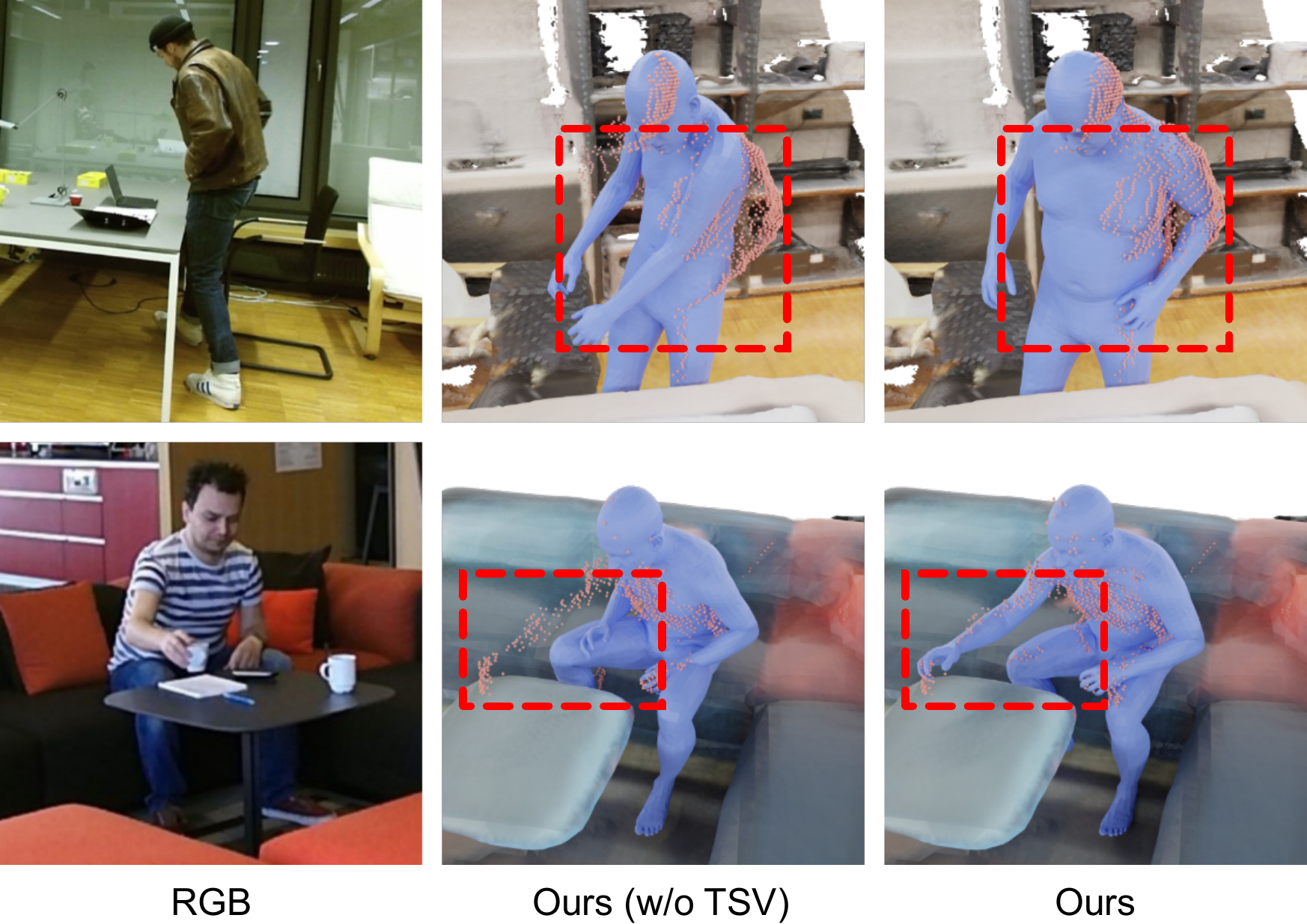}
  \caption{Ablation study: compare our method with the version that does not use truncated shadow volume term. Pink points denote the scanned body point cloud. The difference between results is highlighted using red dashed box.}
  \label{figure:ablation_sv}
\end{figure}

Without the truncated shadow volume term, the reconstructed results do not match with the scanned body point cloud well and the PM score decreases. Figure \ref{figure:ablation_sv} presents two representative examples to demonstrate this effect. In the first example, when our method is applied without the truncated shadow volume term, the reconstructed result displays a mismatch between the left hand and the scanned body point cloud, leading to an unnatural pose. In the second example, although the pose is natural, there is a clear mismatch between the body and the RGB image because the right hand is positioned incorrectly. These examples highlight the important role played by the truncated shadow volume term. By incorporating the truncated shadow volume term, there is a better alignment with the scanned body point cloud, resulting in a visually coherent reconstruction.

\begin{figure}[t]
  \centering
  \includegraphics[width=\linewidth]{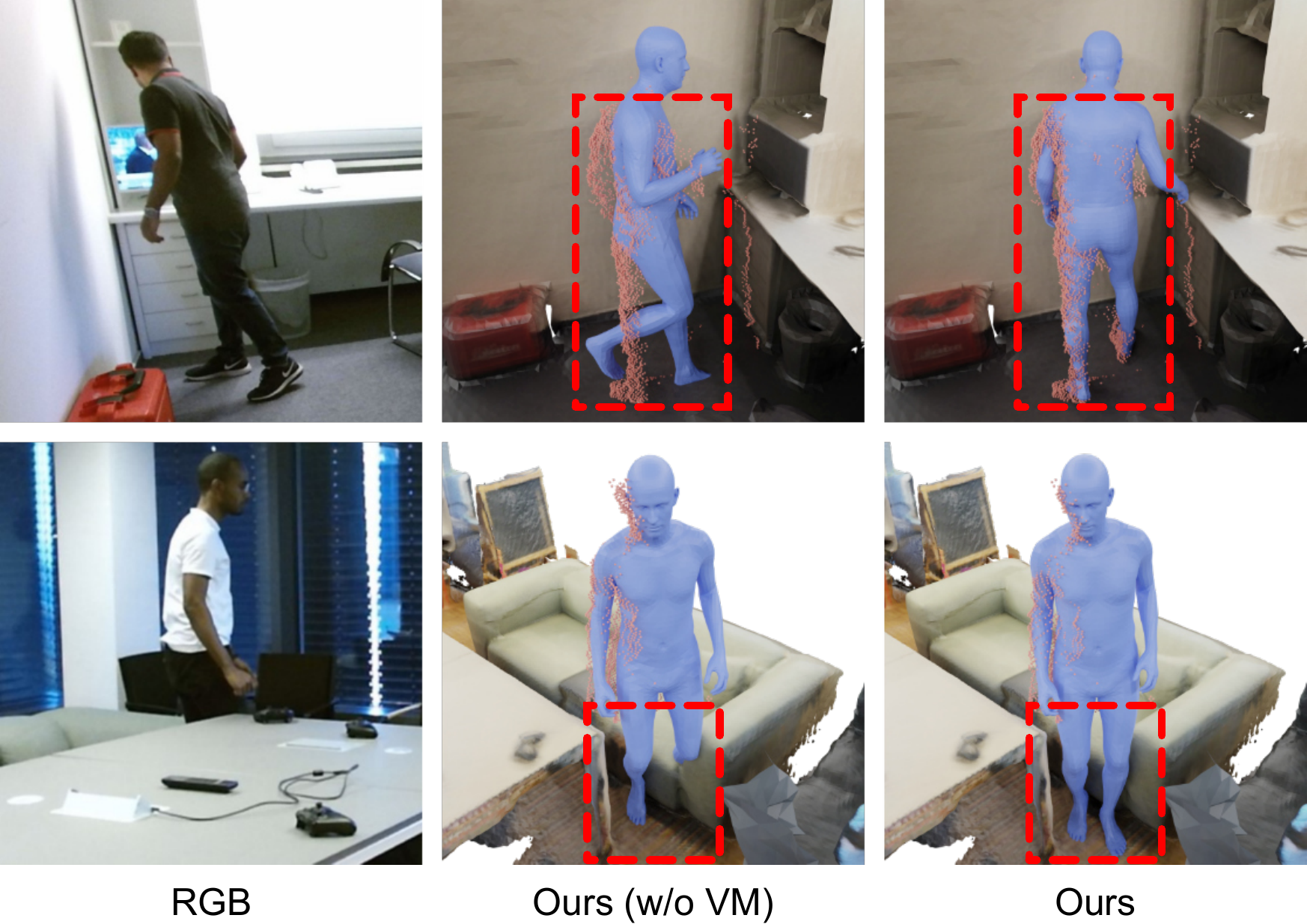}
  \caption{Ablation study: compare our method with the version that uses surface matching instead of volume matching. Pink points denote the scanned body point cloud. The difference between results is highlighted using red dashed box.}
  \label{figure:ablation_vm}
\end{figure}

When the volume matching is replaced with surface matching, \deln{the effectiveness of the free zone term and truncated shadow volume term both decrease}\addn{the effectiveness of both the free zone term and truncated shadow volume term decreases}, and all metrics have a slight decline. Figure \ref{figure:ablation_vm} presents two representative examples to demonstrate this effect. In the first example, when volume matching is replaced with surface matching, the reconstructed result does not match with the scanned body point cloud correctly. In the second example, the human body does not match with the free zone correctly and the left leg penetrates into the sofa. After applying the volume matching algorithm, the reconstructed result is more accurate and plausible. These examples demonstrate the importance of volume matching algorithm. By matching with the confined region using internal points, the free zone and truncated shadow volume can have a more significant effect.

\section{Conclusion and Discussions}

We present a novel method for 3D human body reconstruction from a monocular RGB-D image and the scene mesh. We introduce two schemes to explicitly reduce the solution space based on scene information and prior knowledge. We show that the free zone term can reduce the penetrations and the truncated shadow volume term can increase the matching accuracy. Additionally, a novel volume matching algorithm is proposed to match the human body with the confined region, which yields better performance than surface matching. We also introduce a more comprehensive metric, Volume Non-Collision (VNC), to evaluate the \metricname{} by considering the entire body as a volume. Extensive experiments demonstrate that the proposed method produces more accurate and plausible results compared to other methods, especially in situations with close interactions and serious occlusions.



\noindent\textbf{Limitations and future work}: Our method is deterministic and can only reconstruct one result every time. However, there exists multiple possible results due to occlusion and diversity of pose. In the future, we will explore how to reconstruct multiple possible results which are diverse and plausible. Furthermore, our method assumes a rigid scene, neglecting the deformations that may occur during human-scene interactions in real-world scenarios. To enhance the realism of the reconstructed results, we \deln{will aim to incorporate}\addn{consider incorporating} scene deformations into our approach in future work.

\section*{Acknowledgments}

This work was supported by National Natural Science Foundation of China (62072366), National Key R\&D Program of China (2022YFB3303200), Special Key Projects of Guiding Technological Innovation in Shaanxi Province: 2021QFY01-03 and Key Plan of New Technology \& New Business in Tang Dou Hospital: XJSXYW2021001.


\bibliographystyle{eg-alpha-doi}
\bibliography{sections/reference}

\newcommand{\etalchar}[1]{$^{#1}$}
\begin{thebibliography}{\uppercase{CPMAMN22}}

\bibitem[BKL{\etalchar{*}}16]{SMPLify}
\textsc{Bogo F., Kanazawa A., Lassner C., Gehler P., Romero J., Black M.~J.}:
\newblock Keep it smpl: Automatic estimation of 3d human pose and shape from a single image.
\newblock In \emph{Computer Vision--ECCV 2016: 14th European Conference, Amsterdam, The Netherlands, October 11-14, 2016, Proceedings, Part V 14} (2016), Springer, pp.~561--578.

\bibitem[CHS{\etalchar{*}}21]{OpenPose}
\textsc{Cao Z., Hidalgo G., Simon T., Wei S.-E., Sheikh Y.}:
\newblock Openpose: Realtime multi-person 2d pose estimation using part affinity fields.
\newblock \emph{IEEE Transactions on Pattern Analysis and Machine Intelligence 43}, 1 (2021), 172--186.
\newblock \href {https://doi.org/10.1109/TPAMI.2019.2929257} {\path{doi:10.1109/TPAMI.2019.2929257}}.

\bibitem[CMPM20]{UDF}
\textsc{Chibane J., Mir A., Pons-Moll G.}:
\newblock Neural unsigned distance fields for implicit function learning.
\newblock In \emph{Proceedings of the 34th International Conference on Neural Information Processing Systems} (Red Hook, NY, USA, 2020), NIPS'20, Curran Associates Inc.

\bibitem[CPMAMN22]{LVD}
\textsc{Corona E., Pons-Moll G., Aleny\`{a} G., Moreno-Noguer F.}:
\newblock Learned vertex descent: A new direction for 3d human model fitting.
\newblock In \emph{Computer Vision – ECCV 2022: 17th European Conference, Tel Aviv, Israel, October 23–27, 2022, Proceedings, Part II} (Berlin, Heidelberg, 2022), Springer-Verlag, p.~146–165.
\newblock URL: \url{https://doi.org/10.1007/978-3-031-20086-1_9}, \href {https://doi.org/10.1007/978-3-031-20086-1_9} {\path{doi:10.1007/978-3-031-20086-1_9}}.

\bibitem[CPSA17]{DeepLabV3}
\textsc{Chen L.-C., Papandreou G., Schroff F., Adam H.}:
\newblock Rethinking atrous convolution for semantic image segmentation.
\newblock \emph{arXiv preprint arXiv:1706.05587} (2017).

\bibitem[Cro77]{ShadowVolume}
\textsc{Crow F.~C.}:
\newblock Shadow algorithms for computer graphics.
\newblock \emph{SIGGRAPH Comput. Graph. 11}, 2 (jul 1977), 242–248.
\newblock URL: \url{https://doi.org/10.1145/965141.563901}, \href {https://doi.org/10.1145/965141.563901} {\path{doi:10.1145/965141.563901}}.

\bibitem[GEM87]{GMEF}
\textsc{GEMAN S.}:
\newblock Statistical methods for tomographic image restoration.
\newblock \emph{Bull. Internat. Statist. Inst. 52} (1987), 5--21.

\bibitem[HCTB19]{PROX}
\textsc{Hassan M., Choutas V., Tzionas D., Black M.}:
\newblock Resolving 3d human pose ambiguities with 3d scene constraints.
\newblock In \emph{2019 IEEE/CVF International Conference on Computer Vision (ICCV)} (2019), pp.~2282--2292.
\newblock \href {https://doi.org/10.1109/ICCV.2019.00237} {\path{doi:10.1109/ICCV.2019.00237}}.

\bibitem[HGT{\etalchar{*}}21]{POSA}
\textsc{Hassan M., Ghosh P., Tesch J., Tzionas D., Black M.~J.}:
\newblock Populating 3d scenes by learning human-scene interaction.
\newblock In \emph{2021 IEEE/CVF Conference on Computer Vision and Pattern Recognition (CVPR)} (2021), pp.~14703--14713.
\newblock \href {https://doi.org/10.1109/CVPR46437.2021.01447} {\path{doi:10.1109/CVPR46437.2021.01447}}.

\bibitem[JNV21]{EFT}
\textsc{Joo H., Neverova N., Vedaldi A.}:
\newblock Exemplar fine-tuning for 3d human model fitting towards in-the-wild 3d human pose estimation.
\newblock In \emph{2021 International Conference on 3D Vision (3DV)} (2021), pp.~42--52.
\newblock \href {https://doi.org/10.1109/3DV53792.2021.00015} {\path{doi:10.1109/3DV53792.2021.00015}}.

\bibitem[KB14]{Adam}
\textsc{Kingma D.~P., Ba J.}:
\newblock Adam: A method for stochastic optimization.
\newblock \emph{arXiv preprint arXiv:1412.6980} (2014).

\bibitem[KBJM18]{HMR}
\textsc{Kanazawa A., Black M.~J., Jacobs D.~W., Malik J.}:
\newblock End-to-end recovery of human shape and pose.
\newblock In \emph{2018 IEEE/CVF Conference on Computer Vision and Pattern Recognition} (2018), pp.~7122--7131.
\newblock \href {https://doi.org/10.1109/CVPR.2018.00744} {\path{doi:10.1109/CVPR.2018.00744}}.

\bibitem[KPBD19]{SPIN}
\textsc{Kolotouros N., Pavlakos G., Black M., Daniilidis K.}:
\newblock Learning to reconstruct 3d human pose and shape via model-fitting in the loop.
\newblock In \emph{2019 IEEE/CVF International Conference on Computer Vision (ICCV)} (2019), pp.~2252--2261.
\newblock \href {https://doi.org/10.1109/ICCV.2019.00234} {\path{doi:10.1109/ICCV.2019.00234}}.

\bibitem[KYZ{\etalchar{*}}20]{GraspingField}
\textsc{Karunratanakul K., Yang J., Zhang Y., Black M.~J., Muandet K., Tang S.}:
\newblock Grasping field: Learning implicit representations for human grasps.
\newblock In \emph{2020 International Conference on 3D Vision (3DV)} (2020), pp.~333--344.
\newblock \href {https://doi.org/10.1109/3DV50981.2020.00043} {\path{doi:10.1109/3DV50981.2020.00043}}.

\bibitem[LMR{\etalchar{*}}15]{SMPL}
\textsc{Loper M., Mahmood N., Romero J., Pons-Moll G., Black M.~J.}:
\newblock Smpl: A skinned multi-person linear model.
\newblock \emph{ACM Trans. Graph. 34}, 6 (nov 2015).
\newblock URL: \url{https://doi.org/10.1145/2816795.2818013}, \href {https://doi.org/10.1145/2816795.2818013} {\path{doi:10.1145/2816795.2818013}}.

\bibitem[LSS{\etalchar{*}}22]{MocapDeform}
\textsc{Li Z., Shimada S., Schiele B., Theobalt C., Golyanik V.}:
\newblock Mocapdeform: Monocular 3d human motion capture in deformable scenes.
\newblock In \emph{2022 International Conference on 3D Vision (3DV)} (2022), pp.~1--11.
\newblock \href {https://doi.org/10.1109/3DV57658.2022.00013} {\path{doi:10.1109/3DV57658.2022.00013}}.

\bibitem[LWL21]{METRO}
\textsc{Lin K., Wang L., Liu Z.}:
\newblock End-to-end human pose and mesh reconstruction with transformers.
\newblock In \emph{2021 IEEE/CVF Conference on Computer Vision and Pattern Recognition (CVPR)} (2021), pp.~1954--1963.
\newblock \href {https://doi.org/10.1109/CVPR46437.2021.00199} {\path{doi:10.1109/CVPR46437.2021.00199}}.

\bibitem[MGT{\etalchar{*}}19]{AMASS}
\textsc{Mahmood N., Ghorbani N., Troje N.~F., Pons-Moll G., Black M.}:
\newblock Amass: Archive of motion capture as surface shapes.
\newblock In \emph{2019 IEEE/CVF International Conference on Computer Vision (ICCV)} (2019), pp.~5441--5450.
\newblock \href {https://doi.org/10.1109/ICCV.2019.00554} {\path{doi:10.1109/ICCV.2019.00554}}.

\bibitem[MON{\etalchar{*}}19]{OccNet}
\textsc{Mescheder L., Oechsle M., Niemeyer M., Nowozin S., Geiger A.}:
\newblock Occupancy networks: Learning 3d reconstruction in function space.
\newblock In \emph{2019 IEEE/CVF Conference on Computer Vision and Pattern Recognition (CVPR)} (2019), pp.~4455--4465.
\newblock \href {https://doi.org/10.1109/CVPR.2019.00459} {\path{doi:10.1109/CVPR.2019.00459}}.

\bibitem[PCG{\etalchar{*}}19]{SMPL-X}
\textsc{Pavlakos G., Choutas V., Ghorbani N., Bolkart T., Osman A.~A., Tzionas D., Black M.~J.}:
\newblock Expressive body capture: 3d hands, face, and body from a single image.
\newblock In \emph{2019 IEEE/CVF Conference on Computer Vision and Pattern Recognition (CVPR)} (2019), pp.~10967--10977.
\newblock \href {https://doi.org/10.1109/CVPR.2019.01123} {\path{doi:10.1109/CVPR.2019.01123}}.

\bibitem[PFS{\etalchar{*}}19]{DeepSDF}
\textsc{Park J.~J., Florence P., Straub J., Newcombe R., Lovegrove S.}:
\newblock Deepsdf: Learning continuous signed distance functions for shape representation.
\newblock In \emph{2019 IEEE/CVF Conference on Computer Vision and Pattern Recognition (CVPR)} (2019), pp.~165--174.
\newblock \href {https://doi.org/10.1109/CVPR.2019.00025} {\path{doi:10.1109/CVPR.2019.00025}}.

\bibitem[PIT{\etalchar{*}}16]{DeepCut}
\textsc{Pishchulin L., Insafutdinov E., Tang S., Andres B., Andriluka M., Gehler P., Schiele B.}:
\newblock Deepcut: Joint subset partition and labeling for multi person pose estimation.
\newblock In \emph{2016 IEEE Conference on Computer Vision and Pattern Recognition (CVPR)} (2016), pp.~4929--4937.
\newblock \href {https://doi.org/10.1109/CVPR.2016.533} {\path{doi:10.1109/CVPR.2016.533}}.

\bibitem[PLR19]{BPS}
\textsc{Prokudin S., Lassner C., Romero J.}:
\newblock Efficient learning on point clouds with basis point sets.
\newblock In \emph{2019 IEEE/CVF International Conference on Computer Vision Workshop (ICCVW)} (2019), pp.~3072--3081.
\newblock \href {https://doi.org/10.1109/ICCVW.2019.00370} {\path{doi:10.1109/ICCVW.2019.00370}}.

\bibitem[SCH{\etalchar{*}}16]{PiGraphs}
\textsc{Savva M., Chang A.~X., Hanrahan P., Fisher M., Nie\ss{}ner M.}:
\newblock Pigraphs: Learning interaction snapshots from observations.
\newblock \emph{ACM Trans. Graph. 35}, 4 (jul 2016).
\newblock URL: \url{https://doi.org/10.1145/2897824.2925867}, \href {https://doi.org/10.1145/2897824.2925867} {\path{doi:10.1145/2897824.2925867}}.

\bibitem[SHX{\etalchar{*}}22]{RaG}
\textsc{She Q., Hu R., Xu J., Liu M., Xu K., Huang H.}:
\newblock Learning high-dof reaching-and-grasping via dynamic representation of gripper-object interaction.
\newblock \emph{ACM Trans. Graph. 41}, 4 (jul 2022).
\newblock URL: \url{https://doi.org/10.1145/3528223.3530091}, \href {https://doi.org/10.1145/3528223.3530091} {\path{doi:10.1145/3528223.3530091}}.

\bibitem[SYL15]{cVAE}
\textsc{Sohn K., Yan X., Lee H.}:
\newblock Learning structured output representation using deep conditional generative models.
\newblock In \emph{Proceedings of the 28th International Conference on Neural Information Processing Systems - Volume 2} (Cambridge, MA, USA, 2015), NIPS'15, MIT Press, p.~3483–3491.

\bibitem[TZLW22]{HMRSurvey}
\textsc{Tian Y., Zhang H., Liu Y., Wang L.}:
\newblock Recovering 3d human mesh from monocular images: A survey.
\newblock \emph{arXiv preprint arXiv:2203.01923} (2022).

\bibitem[WCR{\etalchar{*}}19]{GPA}
\textsc{Wang Z., Chen L., Rathore S., Shin D., Fowlkes C.}:
\newblock Geometric pose affordance: 3d human pose with scene constraints.
\newblock \emph{arXiv preprint arXiv:1905.07718} (2019).

\bibitem[XBPM22]{CHORE}
\textsc{Xie X., Bhatnagar B.~L., Pons-Moll G.}:
\newblock Chore: Contact, human and object reconstruction from a single rgb image.
\newblock In \emph{Computer Vision – ECCV 2022: 17th European Conference, Tel Aviv, Israel, October 23–27, 2022, Proceedings, Part II} (Berlin, Heidelberg, 2022), Springer-Verlag, p.~125–145.
\newblock URL: \url{https://doi.org/10.1007/978-3-031-20086-1_8}, \href {https://doi.org/10.1007/978-3-031-20086-1_8} {\path{doi:10.1007/978-3-031-20086-1_8}}.

\bibitem[ZHN{\etalchar{*}}20]{PSI}
\textsc{Zhang Y., Hassan M., Neumann H., Black M.~J., Tang S.}:
\newblock Generating 3d people in scenes without people.
\newblock In \emph{2020 IEEE/CVF Conference on Computer Vision and Pattern Recognition (CVPR)} (2020), pp.~6193--6203.
\newblock \href {https://doi.org/10.1109/CVPR42600.2020.00623} {\path{doi:10.1109/CVPR42600.2020.00623}}.

\bibitem[ZMZ{\etalchar{*}}22]{EgoBody}
\textsc{Zhang S., Ma Q., Zhang Y., Qian Z., Kwon T., Pollefeys M., Bogo F., Tang S.}:
\newblock Egobody: Human body shape and motion of interacting people from head-mounted devices.
\newblock In \emph{Computer Vision--ECCV 2022: 17th European Conference, Tel Aviv, Israel, October 23--27, 2022, Proceedings, Part VI} (2022), Springer, pp.~180--200.

\bibitem[ZWK14]{IBS}
\textsc{Zhao X., Wang H., Komura T.}:
\newblock Indexing 3d scenes using the interaction bisector surface.
\newblock \emph{ACM Trans. Graph. 33}, 3 (jun 2014).
\newblock URL: \url{https://doi.org/10.1145/2574860}, \href {https://doi.org/10.1145/2574860} {\path{doi:10.1145/2574860}}.

\bibitem[ZZB{\etalchar{*}}21]{LEMO}
\textsc{Zhang S., Zhang Y., Bogo F., Pollefeys M., Tang S.}:
\newblock Learning motion priors for 4d human body capture in 3d scenes.
\newblock In \emph{2021 IEEE/CVF International Conference on Computer Vision (ICCV)} (2021), pp.~11323--11333.
\newblock \href {https://doi.org/10.1109/ICCV48922.2021.01115} {\path{doi:10.1109/ICCV48922.2021.01115}}.

\bibitem[ZZM{\etalchar{*}}20]{PLACE}
\textsc{Zhang S., Zhang Y., Ma Q., Black M.~J., Tang S.}:
\newblock Place: Proximity learning of articulation and contact in 3d environments.
\newblock In \emph{2020 International Conference on 3D Vision (3DV)} (2020), pp.~642--651.
\newblock \href {https://doi.org/10.1109/3DV50981.2020.00074} {\path{doi:10.1109/3DV50981.2020.00074}}.

\end{thebibliography}




\section*{Appendix A: Computational Efficiency}

Running time for all configurations is listed in Table \ref{table:time}. We can see that our method achieves better performance than PROX-D with less running time. Besides, volume matching outperforms surface matching both in time efficiency and performance. We also calculated the inference time of the free zone network and found it costs about 0.5 seconds, which means that the bottleneck lies in the optimization process. 

\begin{table}[htb]
    \centering
    \small
    \caption{Running time for all configurations.}
    \resizebox{\linewidth}{!}{
    \begin{tabular}{l|ccccc|c}
    \toprule
        & $\pterm$ & $\cterm$ & $\fzterm$ & $\svterm$ & VM & Running time / s \\
    \midrule
        SMPLify-D & & & & & & 36.63 \\
        PROX-D & \checkmark & \checkmark & & & & 52.13 \\
        Ours (w/o FZ) & & \checkmark & & \checkmark & \checkmark & 47.79 \\
        Ours (w/o TSV) & & \checkmark & \checkmark & & \checkmark & 48.87 \\
        Ours (w/o VM) & & \checkmark & \checkmark & \checkmark & & 51.49 \\
        \textbf{Ours} & & \checkmark & \checkmark & \checkmark & \checkmark & \textbf{50.76} \\
    \bottomrule
    \end{tabular}}
    \label{table:time}
\end{table}

\section*{Appendix B: Performance on Difficult Data}

\begin{figure}[t]
  \centering
  \includegraphics[width=\linewidth]{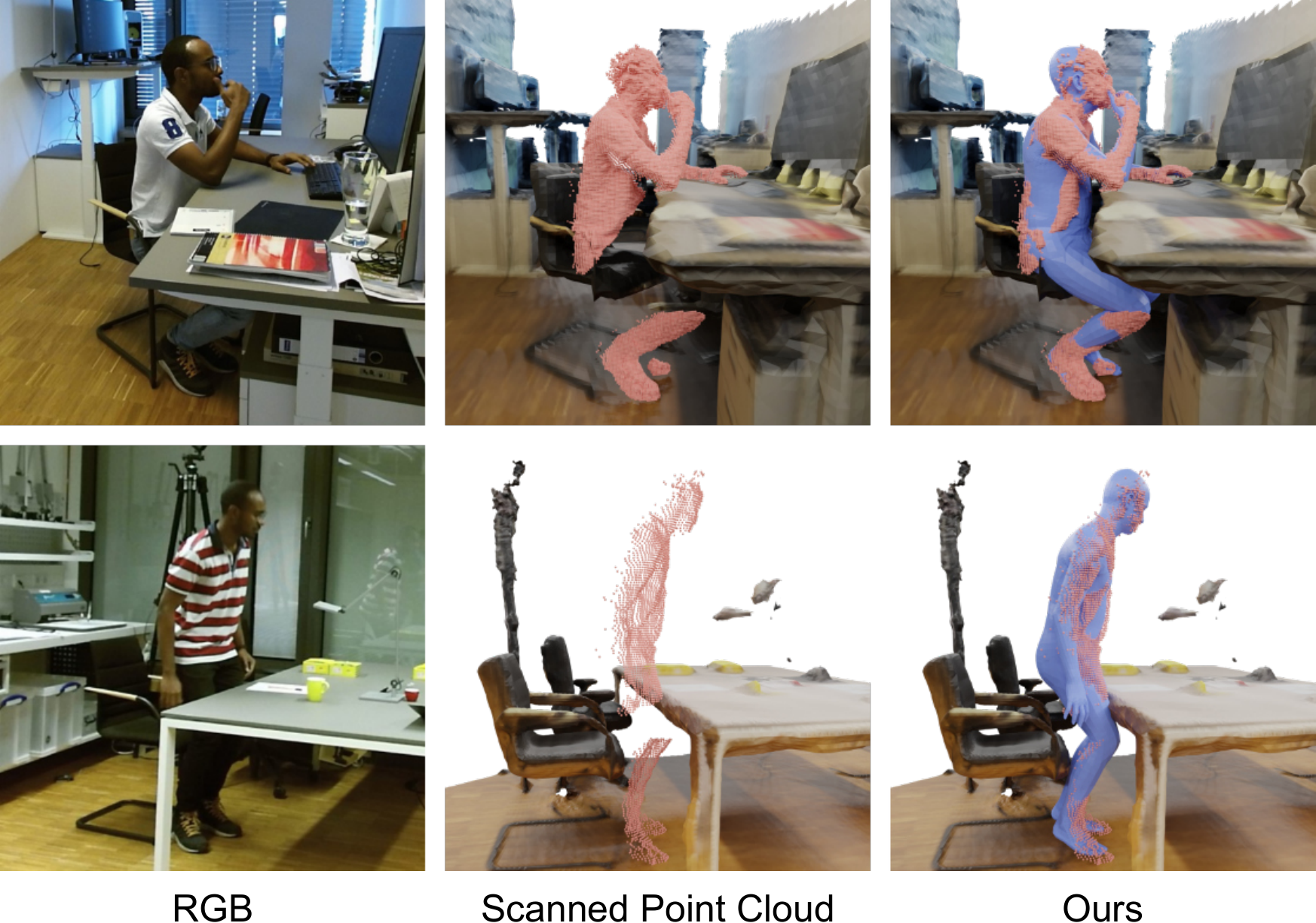}
  \caption{Additional results on cases occluded in the middle.}
  \label{figure:sup_occlusion}
\end{figure}

\textbf{Cases Occluded in the Middle}: In some situations, the human body might be divided into multiple parts due to occlusions. However, we do not perform any special treatments for these cases because we match the body with the confined region holistically. We find it does not affect the performance of our method. Figure \ref{figure:sup_occlusion} presents two representative examples of these cases. In these examples, the scanned body point cloud is divided into the upper part and lower part by the table. Our method can still reconstruct the accurate and plausible pose where the body part matches with the corresponding scanned point cloud regions correctly. 

\begin{figure}[t]
  \centering
  \includegraphics[width=\linewidth]{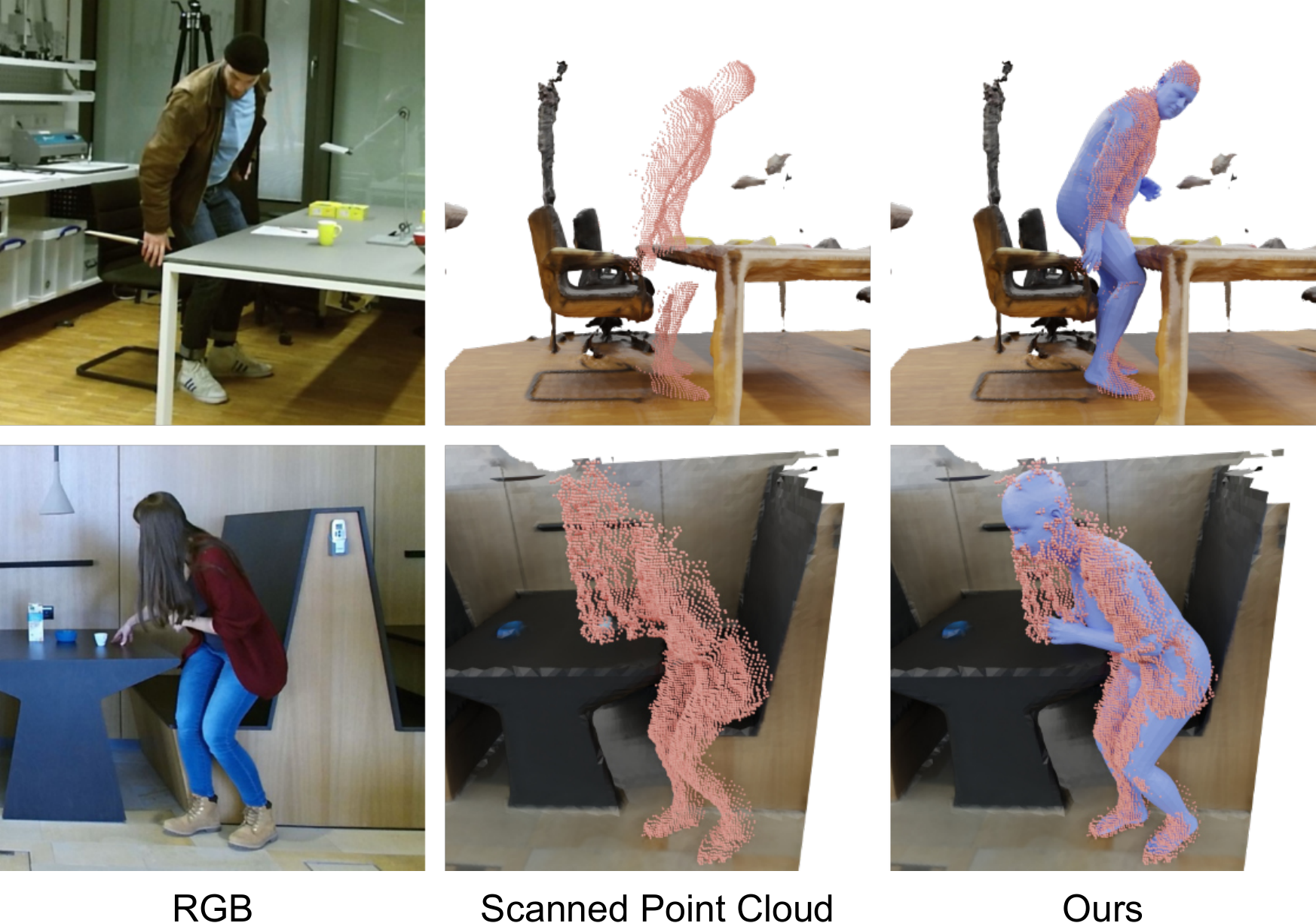}
  \caption{Additional results on cases with loose and large clothes.}
  \label{figure:sup_large}
\end{figure}

\noindent \textbf{Cases with Loose and Large Clothes}: We selected some samples where the human subjects are wearing jackets or coats and observed that our method can still perform well in these cases. Figure \ref{figure:sup_large} presents two representative examples of these cases. When the human is wearing a jacket or a coat, our method can reconstruct the correct pose while aligning the shape with the scanned point cloud. It is worth noting that the body shape might seem a little fatter than the real body shape because SMPL-X model is trained on data of humans wearing tight clothes. To better handle such cases, using a clothed human body model might be a more suitable solution.

\section*{Appendix C: Failure Cases}

\textbf{Error Brought by Discretization}: Discretization might bring some errors, especially in non-convex parts like hands, because the interpolated points can be wrongly located outside the body surface and cause inaccurate matching. In non-convex parts, the two vertices in the same interpolation vertex pair might come from different convex blocks, so the interpolated points might go out of the body, causing the part to lose the original pose and shape. Take the hand as an example, interpolated points of two vertices from different fingers might appear in the finger gaps, making it hard to match the hand with the scanned point cloud accurately. However, these errors can be ignored in our system because the non-convex parts only occupy a small ratio of the whole body. 

\begin{figure}[t]
  \centering
  \includegraphics[width=\linewidth]{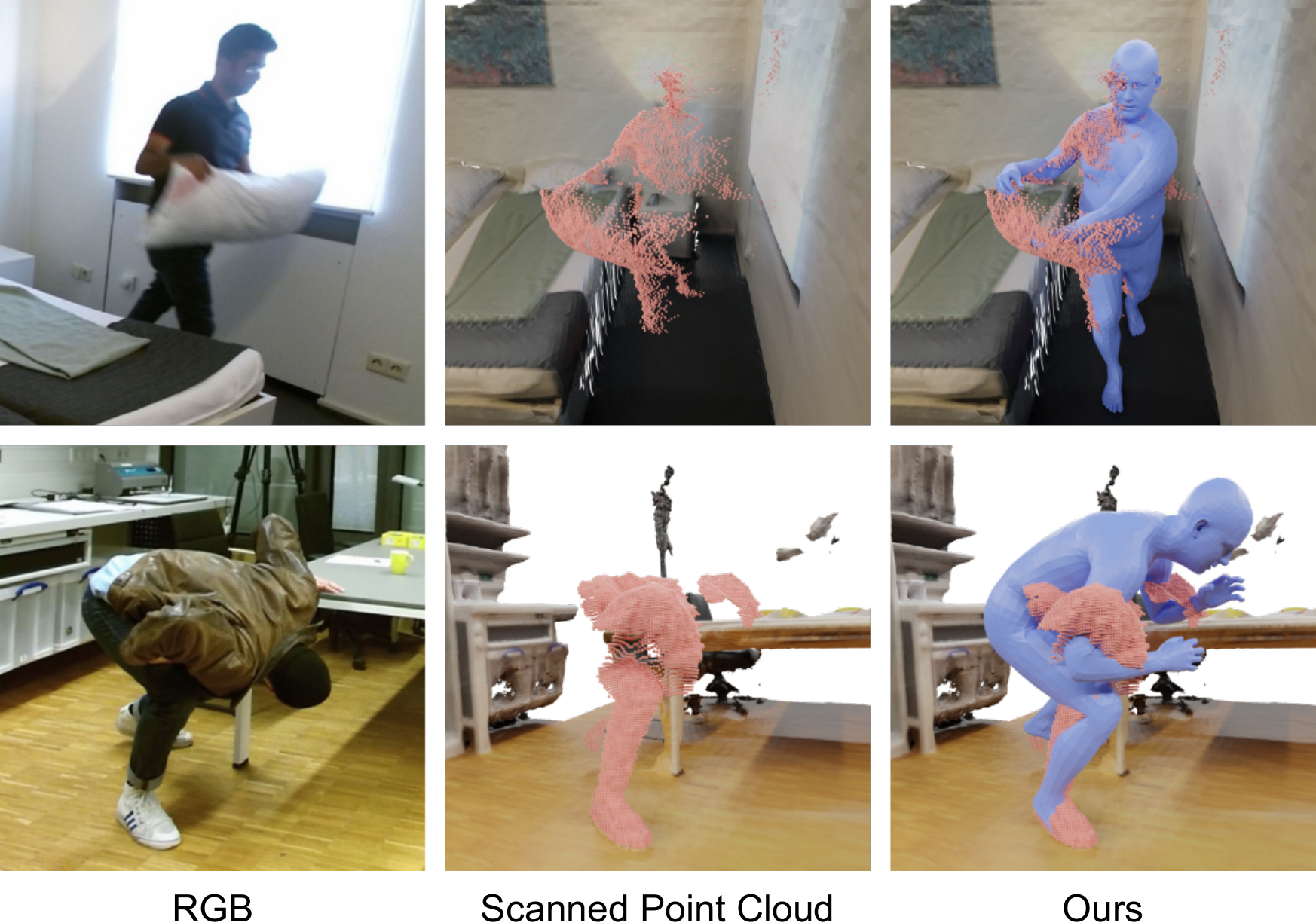}
  \caption{Additional results of failure cases.}
  \label{figure:sup_failure}
\end{figure}

\noindent \textbf{Cases with Poor Scanned Point Cloud}: Our method might fail when the quality of the scanned body point cloud is poor, especially when the body segmentation mask is inaccurate or there exists complex self-occlusion. Figure \ref{figure:sup_failure} presents two failure cases of our method. In the first example, when the human is holding a pillow, and the pillow is wrongly recognized as part of the body, our method produces an incorrect pose where the left hand penetrates into the pillow. In the second example, although the human is not occluded by the scene, the scanned point cloud is incomplete due to the wrong body segmentation mask and complex self-occlusion. Our method fails to reconstruct the accurate pose that matches with the input.

\end{document}